\pdfoutput=1

\documentclass[11pt]{article}

\usepackage[final]{acl}

\usepackage{times}
\usepackage{latexsym}

\usepackage[T1]{fontenc}

\usepackage[utf8]{inputenc}

\usepackage{microtype}

\usepackage{inconsolata}

\usepackage{amsfonts}
\usepackage{amsmath}
\usepackage{graphicx}
\usepackage{caption}
\usepackage{subcaption}

\usepackage{authblk}

\title{Characterizing Similarities and Divergences in Conversational Tones in Humans and LLMs by Sampling with People}

\author[1, 2]{\textbf{Dun-Ming Huang}}
\author[2]{\textbf{Pol van Rijn}}
\author[3]{\textbf{Ilia Sucholutsky}}
\author[4]{\textbf{Raja Marjieh}}
\author[2]{\textbf{Nori Jacoby}}
\affil[1]{Department of Electrical Engineering and Computer Sciences, University of California, Berkeley}
\affil[2]{Computational Auditory Perception Group, Max Planck Institute for Empirical Aesthetics}
\affil[3]{Department of Computer Science, Princeton University}
\affil[4]{Department of Psychology, Princeton University}

\begin{document}
\maketitle

\begin{figure*}[ht!]
    \centering
    \includegraphics[width=\textwidth]{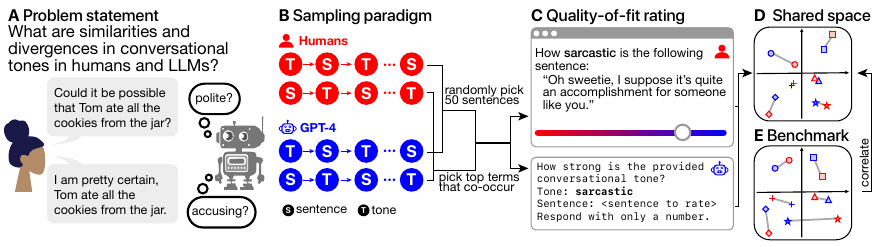}
    \caption{Summary of our approach. \textbf{A}: Problem statement. \textbf{B}: The Sampling with People paradigm that aims to collect a representative sample of conversational tones and sentences. \textbf{C}: A quality-of-fit rating procedure that allows us to obtain vector representations of conversational tones with respect to their usage context. \textbf{D}: A geometric representation of the shared embedding space across elicited domains (human, GPT). \textbf{E}: As an application of our obtained data, we benchmark a selection of popular unsupervised cross-domain alignment methods.}
    \label{fig:schematics-overview}
\end{figure*}

\begin{abstract} 
Conversational tones -- the manners and attitudes in which speakers communicate -- are essential to effective communication. Amidst the increasing popularization of Large Language Models (LLMs) over recent years, it becomes necessary to characterize the divergences in their conversational tones relative to humans. However, existing investigations of conversational modalities rely on pre-existing taxonomies or text corpora, which suffer from experimenter bias and may not be representative of real-world distributions for the studies' psycholinguistic domains. Inspired by methods from cognitive science, we propose an iterative method for simultaneously eliciting conversational tones and sentences, where participants alternate between two tasks: (1) one participant identifies the tone of a given sentence and (2) a different participant generates a sentence based on that tone. We run 100 iterations of this process with human participants and GPT-4, then obtain a dataset of sentences and frequent conversational tones. In an additional experiment, humans and GPT-4 annotated all sentences with all tones. With data from 1,339 human participants, 33,370 human judgments, and 29,900 GPT-4 queries, we show how our approach can be used to create an interpretable geometric representation of relations between conversational tones in humans and GPT-4. This work demonstrates how combining ideas from machine learning and cognitive science can address challenges in human-computer interactions.

\end{abstract}

\section{Introduction} \label{sec:introduction}
Conversational tones, the manner and attitude in which a speaker communicates, is essential to human communication~\citep{YEOMANS2022293,SAIEWITZ201833}. Effective communication relies on people's understanding of conversational patterns and tones, and their ability to promptly react to them~\citep{KREUZ1993239,238ccfea367c4df295cebb4f3e89f586}. Inability to do so results in the content of conversation being ``lost-in-translation'' between speakers of different languages and cultures~\citep{Yusifova2018ThreeLO}. Notably, while traditionally the study of conversational tones involved only humans, the increasing prevalence of Large Language Models (LLMs) in everyday decision-making, especially their conversational (``Chat'') variants, renders the study of conversational tones in LLMs necessary for human alignment~\citep{ouyang2022training, rudolph2023war, sucholutsky2023getting, marjieh2023large}. Developing tools for effectively characterizing conversational tones in humans and LLMs is hence essential for the development of human-centered AI, human-computer interaction research, and cognitive sciences (Figure \ref{fig:schematics-overview}A).

\textbf{\textit{Background: Conversation Research.}} In conversation research, some literature engages explicitly with the composition and usage of conversational attitudes and linguistic markers \citep{VonFintel2006-VONMAL, YEOMANS2022293, JakobsonRoman1960CSLa}. A wider array of literature uses conversational analysis to investigate other dynamics that can affect the content of conversation, such as turn-taking~\citep{238ccfea367c4df295cebb4f3e89f586} and face-saving~\citep{savefaces, facework}, as well as other cross-cultural semantic differences that can lead to different behavior within the same conversational tone, such as refusal~\citep{CHANG2009477}, shame~\citep{shame2018}, and politeness~\citep{Alem_n_Carre_n_2021, Ogiermann+2009+189+216}. On the other hand, the introduction of Large Language Models, especially its chatbot applications, brings attention to the alignment of conversational behavior in LLMs with that of human ideals~\citep{ouyang2022training, rafailov2023direct}, which can potentially contribute to the alignment of LLMs' perception and production of conversational tones with those of humans. Being able to effectively fine-tune LLMs also creates new opportunities to generate text style-transfer corpora that specify sentences with specific conversational tones, such as politeness~\citep{wang2022pay} and formality~\citep{deRivero2023, wang-etal-2019-harnessing}.

\textbf{\textit{Challenge: biased apriori taxonomy.}} However, the domain of conversational tones is, like emotion \citep{schiller2023human,Lindquist2022, athanasiadou1998speaking} and color \citep{Berlin_Kay_1969}, an instance of grounded semantics~\citep{tannen1984pragmatics,doi:10.2753/IMO0020-8825430404}. While all participants observe the same stimulus (e.g., an emotional recording, a solid color, or in our case a sentence), people may use different words or labels to describe it (in our case conversational tones such as ``polite'', ``excited'', and ``grateful'') which makes it difficult to study grounded semantics at scale and especially across multiple languages or cultures. One challenge is that studying grounded semantics often involves adopting a predefined taxonomy, typically sourced from previous studies and curated by investigators (e.g.; colors~\citep{adams1973, wang2016colordrugeffects}; facial emotion~\citep{Ekman1992AnAF}; musical emotion perception~\citep{juslin2008emotional, Palmer2013-nf}, concepts such as animal terms~\citep{10.1162/opmi_a_00072}; sentiment of news items~\citep{Rozado2022} prosody~\citep{sauter2010, Busso2008}. Notably, many machine learning datasets also suffer from the same limitation of using a predefined list of using a predefined list of stimuli that can be outdated or unrepresentative of the correspond modality. Examples of such datasets span through realms of: object images~\citep{imagenet, Krizhevsky2009LearningML}, visual scenes~\citep{zhou2017places}, sounds~\citep{audioset}, video and its categorizations~\citep{kay2017kinetics}, facial expressions~\citep{goodfellow2013challenges}. This strategy is prone to researcher bias, potentially skewing the findings away from an accurate representation of labels as they occur in the real world and within a given culture~\citep{shame2018, henrich2010weirdest}.

\textbf{\textit{Challenge: biased stimulus set.}} Another challenge that almost all studies faced when studying grounded semantics is that they may use a constrained set of stimuli to be annotated (e.g. emotion~\citep{Cowen2017SelfreportC2, Cowen2019, Cowen2020, Cowen2020WhatTF, CowenVocal}; object recognition and similarities~\citep{GIFFORD2022119754, Hebart2019-tb}; word-associations~\citep{dedeyneetal23, DeDeyne2019}; musical perception~\citep{inbook}; facial expression~\citep{khaireddin2021facial, Lin2021}; prosody~\citep{ELAYADI2011572, Aibo2008}).
This introduces researcher bias, as curating the stimuli may influence the elicited labels, which we outline using the following example. Imagine an experiment where a particular semantic term can be associated with some class of objects (e.g., the term ``red'' can be used to describe red fruits). If the object class is not included in the predetermined list of objects (e.g., red fruits are not included in the list of objects), then the elicited terms will not include this association (we will conclude that ``red'' does not describe fruits), and it will be missing from the resulting semantic network. Bias in object selection can also occur in more subtle ways where a skew in the distribution of selected objects also skews the distribution of elicited terms, potentially even amplifying the initial bias. Furthermore, a large body of cross-cultural researchers suggests that studies should not impose a terminology inherited from the experimenter or even from one group of studied agents (e.g., English speakers) on another agent or group of agents \citep[e.g., Speakers of another language or demographics;][]{BLASI20221153,barrett2020towards,henrich2010weirdest}. 

\textbf{\textit{Additional challenges.}} Finally, while some studies advocate for the exclusive use of large textual corpora and the extraction of semantic descriptors via data mining \citep{Thompson2020CulturalIO}, this indirect approach raises concerns about its ability to accurately represent the nuances of conversational tones as experienced in everyday human interactions. It is also difficult to rigorously compare humans and LLMs using such corpora because these same textual corpora are also the basis for LLM training.
Notably, ~\cite{Thompson2020CulturalIO} also studies the problem of aligning semantic networks of different individuals or groups in the context of cross-linguistic and cross-cultural comparisons.
It turns out that this is a key part of the machine learning problem of automatic translation~\citep{10.1162/jocn_a_01000, LIU202173}. Recent research has focused on aligning semantics in humans with Large Language Models~\citep{sucholutsky2023getting, atari_xue_park_blasi_henrich_2023}, with significant applications to designing human-computer interfaces~\citep{Hou2024C2IdeasSC} and AI safety.

\textbf{\textit{Our approach.}} In light of these challenges, we propose a method that enables the characterization of conversational tones and their taxonomies in any target human population as well as LLMs, based on a human-in-the-loop Sampling with People (SP) technique~\citep{SANBORN201063, griffiths2005bayesian,harrison2020gibbs} (Figure \ref{fig:schematics-overview}).
Specifically, we propose an iterative procedure in which humans and LLMs are presented with sentences and are asked to label their conversational tones in an open-ended fashion (Figure \ref{fig:schematics-overview}B). The resulting conversational-tone terms are then presented to a new group of agents who are asked to produce sentences reflecting those conversational tones. This process is then repeated multiple times.
With mathematical formalism, this process instantiates a Gibbs Sampler from the joint distribution of sentences and conversational tones in humans and LLMs~\citep{harrison2020gibbs,griffiths2024estimating}. Given the resulting sample, we derive representative sentences and tone taxonomies of our target population, then have an independent group of human evaluators and LLMs rate the extent to which each tone matched each sentence (Quality-of-fit Rating; Figure \ref{fig:schematics-overview}C). We use these to construct a geometric embedding that can be used to evaluate the alignment between human and LLM conversational tones (Figure \ref{fig:schematics-overview}D).

We show how our approach can be effectively used to reveal divergences in the representation of conversational tones between humans and LLMs. Moreover, we demonstrate how our new dataset and cross-evaluations can be used to benchmark unsupervised cross-domain semantic alignment methods used in existing work, and identify which of these work well for cases in which cross-evaluation is not possible (e.g., in multilingual scenarios; Figure \ref{fig:schematics-overview}E).
Our method can be generalized to many more psycholinguistic modalities (e.g., sentiment, color), languages, and cultures beyond those involved in this paper. We believe it will help advance both human-machine alignment research as well as cross-cultural research.

\section{Detailed Approach}

\subsection{Elicitation via Sampling with People}

The core of our approach is the joint elicitation of a representative sample of conversational tones and sentences from both humans and LLMs.
Specifically, we propose an iterative procedure that is composed of two steps per iteration. Step one, in which humans and LLMs are presented with sentences and are asked to classify their conversational tones in an open-ended fashion (Figure \ref{fig:schematics-overview}B). Step two, the resulting conversational tone descriptors (adjectives) are then presented to a new group of agents, from whom we ask to produce sentences that reflect those conversational tones. This process is then iterated multiple times. Formally, this process instantiates a Gibbs Sampler from the joint distribution of sentences and conversational tones, for any target population we choose to sample from, be it humans and LLMs~\citep{griffiths2024estimating}. Therefore, by constraining the set of human participants to those from a specific cultural group, we expect to obtain a representative sample of psycholinguistic contents from the group of our participant population.
 
Here we draw inspiration from human-in-the-loop elicitation procedures \citep{griffiths2024estimating}. In these methods such as serial reproduction~\citep{xu2010rational,anglada2023large,jacoby2017integer, jacoby2024commonality, langlois2017uncovering,langlois2021serial}, iterated learning \citep{xu2010replicating,griffiths2005bayesian}, MCMCP~\citep{SANBORN201063} and GSP~\citep{harrison2020gibbs, van2022voiceme, van2021exploring, van2024using,marjieh2024timbral,van2024giving}, a Markov Chain is constructed by interspersing human decisions within a sampling chain to characterize latent representations in the human mind (e.g., perceptual prior, or subjective utility).
In our novel approach, Sampling with People (SP), humans are recruited to perform two tasks (Figure \ref{fig:schematics-overview}B): (1) elicit a sentence based on a conversational tone (``S'' task), and (2) annotate a conversational tone of a given sentence (``T'' task).
Under a probability theory framework, ``S'' and ``T'' tasks are essentially sampling operations, respectively from the conditional distribution of a sentence given a conversational tone $p(S | T)$, and the conditional distribution of a tone given a sentence $p(T | S)$.
In practice, we run several parallel sampling chains, and in each trial, a participant is assigned to one chain and performs an ``S'' or ``T'' task as needed. This means that each sampling chain alternates between ``S'' and ``T'' tasks. Importantly, to satisfy the formulation of a Gibbs' Sampler, we design our paradigm to satisfy the Markovian property by constraining each participant to see only the output of the previous iteration~\citep{harrison2020gibbs,griffiths2005bayesian}.

Using SP, we elicited a large database of tones and sentences from humans and LLMs (separately for each). We elicited 40 tones and 80 sentences with 955 participants, 90 chains, and 100 iterations each. The list of 40 conversational tones we investigate is the union of the top 24 conversational tones from each instance of SP experiments. We then took 40 random sentences from each of the humans and GPT, forming a balanced corpus of humans and GPT in terms of sentence sources.
We preserve the representativeness of our sample from the internal distribution of sentences $P(S)$ by choosing the random sentences in a uniform sampling fashion.
In the design of this study, we use a shared array of conversational tones to work with a consensus taxonomy, enabling direct comparability between the conversational behavior of humans and GPT-4 when prompted under the same language\footnote{From here on, all references to GPT-4 will be abbreviated as GPT.}.

\subsection{Annotation via Quality-of-fit Rating}
Given the distributions from prior subsection, we have derived representative sentences and tone taxonomies. Then, we have an group of human evaluators independent from prior participants, as well as GPT rate the extent to which each tone matched each sentence (Quality-of-fit Rating; Figure \ref{fig:schematics-overview}B).
We show how our approach can be effectively used to reveal divergences in the representation of conversational tones between humans and LLMs. Specifically, after SP sampling, we collect quality-of-fit ratings of all sentences with all tones (Figure \ref{fig:schematics-overview}C), and compute semantic similarity matrices of different tones.
Then, for two tones $t_i, t_j$, let $\mathcal{R}_{i,j}$ denote the correlation of the two tones across the vector of average ratings of all sentences, such that $t_i$ and $t_j$ are similar if they have similar ratings across all sentences. We can compute such matrix $\mathcal{R}$ in either an intra-domain manner (only using embeddings from either humans or GPT but not both), or a cross-domain manner (where we exploit the shared list of sentences to compute the correlation between all conversational tone embeddings).

\subsection{Geometric Representation of Conversational Tones}
We use the resulting cross-domain similarity matrices to obtain a geometric representation of both human and GPT data within the same space (Figure \ref{fig:schematics-overview}D). We compute the full correlation of all 80 tone embeddings (40 conversational tones, each one with an embedding from human data and another from GPT data), and use Multidimensional Scaling \citep[MDS;][]{carroll1998multidimensional,ANOWAR2021100378} to project them into a shared low-dimensional embedding space. The space thus represents not only the relation between tones within humans but also the way they relate to GPT, especially as the proximity of tones in the MDS Euclidean space corresponds to the proximity of tones in terms of their semantic similarity~\citep{Shepard1980MultidimensionalST}. Therefore, tones that appear closer in the shared space are located nearer in Euclidean space.

\subsection{Application: Benchmarking Semantic Alignment Methods}
To demonstrate the usability of our alignment data, we show how it can be used to benchmark semantic alignment methods as ground truth when cross-annotation is unavailable (Figure \ref{fig:schematics-overview}E) and only intra-domain correlation matrices are used. Specifically, we benchmark the performances of (1) Gromov-Wasserstein Optimal Transport~\citep{grave2018unsupervised, conneau2017word, kawakita2023comparing}, (2) Bilingual Lexicon Induction~\citep{ruder2018discriminative, artetxe2016emnlp, artetxe2017acl, artetxe2018aaai, artetxe2018acl}, and (3) Orthogonal Procrustes~\citep{Schnemann1966AGS,BEAUDUCEL201820}.

\section{Method} \label{sec:methods}

\subsection{Participants}
\textbf{\textit{Human Participants.}} We recruited \textit{N}=1,339 participants for the four human experiments in this study (Appendix Table \ref{tbl:n_participants} provides the number of participants and responses for each experiment). Participants were recruited from the recruiting platform Prolific and provided informed consent under an approved protocol (see Ethics section for further information). Human experiments are implemented using Psynet~\citep{harrison2020gibbs}, a Python package for implementing complex online psychology experiments. 

\textbf{\textit{GPT experiments}}
We used the June 13th, 2023 release of GPT-4. Overall, we ran 29,900 GPT queries across all experiments.

\subsection{General Procedure} 
Each human experiment began with detailed instructions and practice trials. We emulated each of the human experiments with GPT-4 agents that used the very same procedure, using the experiment interface instructions as LLM prompts. The human and GPT-4 experiments were otherwise identical in their design. Appendix \ref{sec:appendix_methods} contains the full instructions/prompts used in our experiments.
All the data of the experiments, code for reproducible human and GPT experiments, and analysis script can be found here: \url{https://github.com/jacobyn/SamplingTonesACL}.
\begin{figure*}[ht!]
    \centering
    \includegraphics[width=\textwidth]{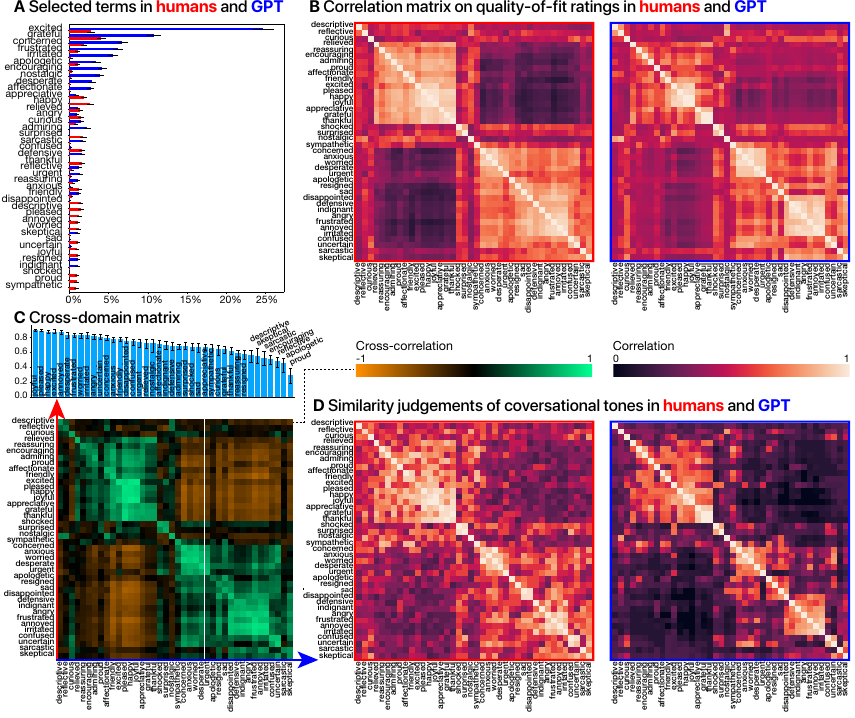}
    \caption{
    Results of Sampling with People and Quality-of-fit Rating paradigms and comparison to similarity judgments paradigm.
    \textbf{A}: Selection of most popular conversational tones from each of human and GPT instances, and their frequencies in respective samples (red for humans, blue for GPT). Error bars represent one standard deviation via bootstrapping.
    \textbf{B}: Correlation matrices of conversational tone quality-of-fit rating embeddings within humans (on the left) and within GPT (on the right).
    \textbf{C}: Cross-domain (Cross-correlation) matrix of human rating embeddings and GPT rating embeddings for conversational tones, and a bar plot showing the correlation between human ratings and GPT rating embeddings for each conversational tone word. Error bars represent one standard deviation via bootstrapping.
    \textbf{D}: Similarity judgment-derived similarity matrices of conversational tone from humans (on the left) and GPT (on the right). See enlarged version of this figure in the Appendix (Figure \ref{fig:dense-large}).
    }
    \label{fig:dense}
    
\end{figure*}

\section{Results} \label{sec:results}

\subsection{Elicitation (Sampling with People)} \label{subsec:sp}
Figure \ref{fig:dense}A shows the histogram of the 24 most popular conversational tones from 4,500 human and 4,500 GPT annotated sentences. We recruited 955 human participants for SP experiments.
The collected histograms were reliable: the split-half reliability computed via bootstrapping\footnote{Throughout the paper, bootstrapping occurs with 5000 repetitions, unless specified otherwise.} was high for both humans and GPT conversational tones (humans: \textit{r} = 0.91, CI = [0.87, 0.93]; GPT: \textit{r} = 0.87 CI = [0.73, 0.94]\footnote{Throughout the paper all CI are reported as 95\%.}). The GPT histogram (Figure \ref{fig:dense}A) was much more concentrated (entropy of 3.10 bits CI = [3.06, 3.15] via bootstrapping) compared to that of humans (entropy of 5.48 bits CI = [5.43, 5.52]).
Overall, there was some similarity between the histograms (\textit{r} = 0.39 \textit{p} = 0.006 CI = [0.3,0.454]), but also significant differences. Specifically, the prominent conversational tones had different weights: in humans the three most prominent conversational tones were ``grateful'' (mean 4.03\% CI = [3.45\%, 4.62\%]), ``excited'' (2.79\% CI = [2.35\%, 3.28\%]), then ``happy'' (2.64\% CI = [2.18\%, 3.1\%]) whereas in GPT they were ``excited'' (mean 24.66\% CI = [23.41\%, 25.91\%]), ``grateful'' (10.79\% CI = [9.89\%, 11.68\%]), then ``concerned'' (6.79\% CI = [6.05\%, 7.52\%]). The results highlight that the elicitation process results in a different distribution of terms used to describe conversational tones. 

\subsection{Annotation via Quality-of-fit Rating} \label{subsec:dr}
To create a detailed semantic embedding based on the given sentences and tones, we conducted a further experiment involving both humans and GPT. In this study, participants evaluated all target sentences using a predefined list of 40 prominent conversational tones that were extracted from the terms elicited using the SP procedure above, ensuring a direct comparison by employing the same tones for both human and GPT assessments. As a result, every sentence is relabelled with all 40 tones regardless of the original tone it was assigned in the SRE step. The sentence set was evenly divided, with one half originating from humans and the other half generated by GPT.

We recruited an additional 275 human participants for these annotations. Participants rate the strength of conversational tones in a sentence using a Likert scale, resulting in 16,000 rating judgments. Likert scales were used because the degree to which a sentence represents a tone may vary and cannot be captured by categorical labels \citep{sucholutsky2023informativeness}.  GPT data underwent the same rating process with GPT agents as raters. We then analyzed the correlation between tones by examining the rating vectors across the 80 sentences elicited from Section \ref{subsec:sp}. Additionally, we find that the majority of sentences are labeled to possess at least one specific conversational tone: all human-originated sentences had at least one conversational tone with an average rating exceeding 2.89, and exceeding a rating of 2 for 95\% of GPT-originated sentences. This would suggest that very few sentences were perceived as completely neutral with respect to all 40 tones.

Figure \ref{fig:dense}B presents the tone-correlation matrices for humans (left) and GPT (right). The matrices show reliable tone-similarity for both humans and GPT (humans: \textit{r} = 0.94 CI [0.91, 0.96];  GPT ratings, \textit{r} = 0.86 CI [0.78, 0.91]). It is visually apparent that there are two clusters of tones which roughly related to the valence of tones and that this structure is preserved in humans and GPT as we found a high correlation between the upper triangle of the two matrices (\textit{r} = 0.81 CI [0.76, 0.85]).

To better understand how humans and GPT are similar and different in the structure of tone-similarity we now compute the cross-domain matrix, namely for each tone we correlated the vector of sentence ratings in humans and in GPT (Figure \ref{fig:dense}C). As expected, given that both humans and GPT show separately the pattern of valance clusters, the cross-domain matrix also showed this pattern. Interestingly, the main diagonal of this matrix shows that the alignment between tone ratings in humans and GPT varies significantly. Some tones such as ``joyful'', ``pleased'', and ``happy'' were highly correlated (\textit{r} = 0.89, 0.89, 0.87 CI = [0.88, 0.91], [0.88, 0.91], [0.86, 0.90], respectively) suggesting that these tones have a similar relation to other tones in both humans and GPT. However, other tones such as ``proud'', ``apologetic'', and ``reflective'' had relatively low alignment (\textit{r} = 0.30, 0.46, 0.49 CI = [0.20, 0.39], [0.34, 0.52], [0.41, 0.54]). 

Finally, our tone similarity calculations were somewhat indirect since we did not compare tone similarity directly but computed it based on the similarities of sentence ratings. 
To address this, we recruited a separate group of 71 participants. Each participant performed 50-60 trials and was asked to provide similarity ratings between two tones on a Likert scale. A similar procedure was applied for GPT. The resulting matrices (\ref{fig:dense}D) indicate that direct similarity judgments also showed the valence block structure, but the similarity matrix appears to be noisier. 
Nevertheless, the data was still reliable: split-half correlations for the upper diagonal or the matrices were humans: \textit{r} = 0.65 CI [0.63, 0.68], and that for GPT's similarity matrix to be mean 0.981 with CI [0.978, 0.987]. The high reliability of GPT partially originates from the monotone response that GPT provides in similarity judgments (unlike humans GPT agents provides the same response again and again for the same input). Importantly, the direct similarity matrices were aligned with the quality-of-fit rating approach (humans: \textit{r} = 0.76 CI = [0.73, 0.78], GPT: \textit{r} = 0.83 CI = [0.81, 0.84]). This is an independent validation that our approach does capture tone similarity structure. It also highlights the difficulty of working with direct similarity \citep{marjieh2023words}.

\begin{figure*}[ht!]
    \centering
    \includegraphics[width=\textwidth]{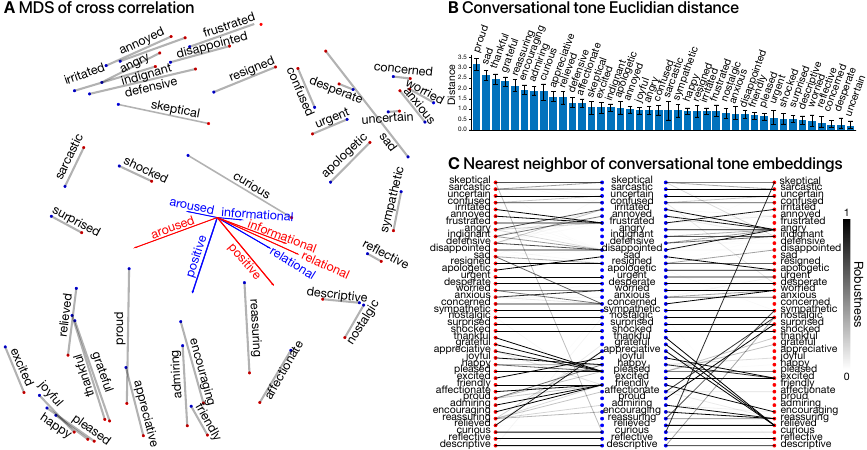}
     \caption{
     Cross-correlation alignment information. Blue points/arrowmarks in \textbf{A} and \textbf{C} represent GPT-originated data, while red represents human-originated instead.
     \textbf{A}: The MDS solution of applied to the combined within/across cross-domain (cross-correlation) matrix as a set of high-dimensional embedding to represent shared space of conversational tones embeddings across humans and GPT. Grey edges connect points representing the same conversational tone word. Arrow marks represent rating-derived dimensions of conversational tones.
     \textbf{B}: A barplot exhibiting the Euclidean distance between pairs of the same conversational tone embeddings in MDS space. Error bars represent one standard deviation via bootstrapping.
     \textbf{C}: A graph showing the nearest neighbor matches of conversational tone embeddings across humans and GPT. To measure robustness in matching, we bootstrapped the process 5000 times. Dark edges represent the frequency of its matching throughout bootstrap processes. See enlarged version of this figure in the Appendix (Figure \ref{fig:mds-large}).
     }
     \label{fig:mds}
\end{figure*}
\subsection{Conversational Tone Representation (Multidimensional scaling)} \label{sec:result-cca}
In order to understand the organization of tone representations, we applied Multidimensional Scaling (MDS) to the combined within/across correlation matrix, such that each conversational tone possesses a 2-dimensional embedding in the resulting shared MDS space (Figure \ref{fig:mds}A). 

We connected identical tones in GPT and humans with gray lines. Overall, it is visually apparent from Figure \ref{fig:mds}A that the structure of tones in humans and GPT is similar, which is consistent with the analysis of Figure \ref{fig:dense}. However, we also found differences in the proximity of tones in the shared space. The furthest tones away were ``proud'', ``sad'', and ``thankful'', while the closest were ``uncertain'', ``desperate'', and ``conncerned'' (Figure \ref{fig:mds}B).

To better interpret what the dimensions of difference are between humans and GPT and compare our results with respect to previous literature~\citep{YEOMANS2022293,russell1980circumplex}, we performed an additional experiment. 38 participants rated each tone from 1 to 5 on a Likert scale based on four theoretical dimensions: valence, emotional arousal, informational, and relational (overall 800 ratings). We also conducted a similar experiment with GPT (800 ratings). These features are defined in Appendix \ref{sec:appendix-methods-feature-def}. To observe how these ratings relate to the conversational tone embedding' MDS solutions, we projected the average rating over tones to the MDS (see Appendix \ref{sec:appendix-cca} for projection method using linear regression). Each theoretical dimension is represented by an arrow (direction) in MDS space, with the length of the arrow representing its relevance (measured by how much the dimension explained the variance among the points in the MDS).

The theoretical component of conversational tone that seems to explain most of the variance in their MDS solutions is ``positive in valence''  (humans: mean \textit{r} = 0.71 CI = [0.69, 0.87]; GPT: mean \textit{r} = 0.873 CI = [0.872, 0.878]), where the other terms explained significantly less variance (GPT: \textit{r} = 0.45, 0.17, 0.13; humans: \textit{r} = 0.37, 0.54, 0.3 for ``relational'', ``aroused'', ``informational'' respectively). This is consistent with the idea that the main axis of the quadratic shown by MDS solutions corresponds roughly to the positive/negative valence (``joyful'', ``happy'', ``excited'', and ``pleased'' are in the bottom left, whereas ``concerned'', ``worried'', and ``anxious'' are in the top right).  
 
From Figure \ref{fig:mds}A, however, we see that the human and GPT arrowmarks for the same conversational tone may have minor to medium directional differences, such as that for the feature ``positive in valence'' and ``aroused'' (indicated by a large cosine similarity; mean = 0.88 CI = [0.86, 0.94] and mean = 0.46 CI = [0.04, 0.53], respectively). Meanwhile, ``relational'' is consistently aligned  (mean = 0.99 CI = [0.86, 0.99]), while ``informational'' is also strongly aligned (mean = 0.83 CI = [-0.26, 0.89]; see Appendix \ref{sec:appendix-methods-feature-def}). This suggests a deviation between the human and GPT understanding of these features. Furthermore, it validates \citeauthor{YEOMANS2022293}'s chosen features for conversational tone composition, ``informational'' and ``relational'', as an aspect of well-aligned conversational understanding between humans and GPT~\citep{YEOMANS2022293}. These results also allow us to further interpret the MDS of Figure 3A, for example, ``proud'' is located farther away from neutral in the MDS space and stronger in the positive valence direction for humans. Thus, ``proud'' has a more neutral meaning for GPT. This shows how we can use Figure 3A to characterize how tones are conveyed in humans and GPT.

Finally, Figure \ref{fig:mds}C provides a mapping of similar tones across humans and LLMs. For each tone in one domain (humans or GPT), we present its nearest neighbor in another domain. This map is important because it allows us to "translate" conversational toes from humans to GPT and vice versa. As a result, Figure \ref{fig:mds}C can also help summarize distances in Figure \ref{fig:mds}A: the lines in Figure \ref{fig:mds}C represent concepts that are near one another in Figure \ref{fig:mds}A. Interestingly, we found cases where multiple human tones (e.g., ``grateful'', ``joyful'', ``happy'') were collapsed to a single LLM tone (``pleased''), suggesting that these terms were conveyed in a more limited way by GPT. We also found the reverse phenomenon that the GPT tones: ``irritated'', ``annoyed'', and ``disappointed'' collapsed to ``angry'' on the human side. This suggests that both humans and GPT have terms that are represented more broadly in the other group.

\subsection{Application: Ground Truth for Benchmarking Semantic Alignment Methods}
To demonstrate the utility of our data, we demonstrate how it can be used to benchmark semantic alignment methods. Note that in our approach we leverage the fact that we had the same taxonomy of tones for both humans and GPT. In many other use cases, this is not possible. For example, when translating words from one language to another it is impossible to ask speakers of one language to annotate words in a different language. Since our method has this kind of cross-linguistic information available, we use it as a source of ground truth to test methods that do not have access to this kind of information. 
Specifically, we survey the capability of frequently used unsupervised cross-domain alignment paradigms: Gromov-Wasserstein Optimal Transport~\citep{grave2018unsupervised}, Orthogonal Procrustes Transformation~\citep{Schnemann1966AGS}, and Bilingual Lexicon Induction~\citep{ruder2018discriminative}. 
We found that Bilingual Lexicon Induction via latent variable model (abbreviated hereon as BLI) is the best-performing approach. It recovers well the quality-of-fit rating similarity matrix entries (\textit{r} = 0.81, CI = [0.81, 0.81]), followed by Orthogonal Procrustes Transformation (\textit{r} = 0.56 CI = [0.54,0.58]) and Gromov-Wasserstein Optimal Transport (\textit{r} = 0.54 CI = [0.53,0.58]). We found similar results for the recovery of k-Nearest-Neighbor structures (see Appendix Table \ref{tbl:benchmark}). This indicates that not only is Bilingual Lexicon Induction superior at recovering the cross-domain proximity structure presented in our cross-domain ground truth, but also that it can preserve the intra-domain proximity structure of our embeddings after the alignment procedure. Then, in turn, this performance turns to show the superiority of our method in constructing a dataset that allows for almost-perfect reconstruction of similarity metrics for elements of the psycholinguistic modality.

\section{Discussion}
Using a cognitive science-inspired Sampling with People paradigm, we elicited tones and sentences for both humans and GPT creating a shared dataset of sentences and tones. Then, via quality-of-fit ratings, we further showed that the degree of alignment for different tones in humans and GPT varies. Tone alignment was high for tones such as ``joyful'' and ``pleased'' while it was significantly lower for tones such as ``proud'' and ``apologetic''. In a separate experiment, we found that similarity judgments were consistent with the resulting relationship between tones. Next, by projecting the rating vectors to a joint semantic space, we found variability in tone proximity, which is explained most by the well-known theoretical dimension of valence~\citep{russell1980circumplex}. Additionally, we provided a mapping of similar tones across humans and LLMs. We found cases where multiple human tones (e.g., ``grateful'', ``joyful'', ``happy'') were collapsed to a single LLM-proposed tone (``pleased''), as well as in the opposite direction (e.g., GPT's ``irritated'', ``annoyed'', and ``disappointed'' collapsed onto humans' ``angry''), suggesting that both distributions ended up involving a hypernym for conversational tone categories. Finally, we demonstrate how our data can be used to benchmark methods for semantic alignment. We found that Bilingual Lexical Induction surpasses other (geometric) methods, suggesting that it would be appropriate for applications such as machine translation.

Our work opens up multiple avenues for future research. First, the same approach can be easily extended to other domains. For example, speakers of different languages and different cultures, as well as different language models. Second, our dataset can be used as a training signal for better aligning human and LLM conversational tones; for example, via an iterative process of refinements like reinforcement learning~\citep{ji2024ai}. Third, future work could look into whether we can use our alignment maps to predict performance in human-AI communication.  More broadly, this work shows how combining approaches from machine learning and cognitive science provides routes for better understanding and resolving challenges in human-computer interactions.

\section*{Limitations}
There are several technical limitations of our study that are important to highlight. First, we did not test a wide variety of LLMs and LLM parameters, including varying the prompt and temperature of the models. This limits the generalizability of our results, as the robustness of some of our findings may depend on these parameters. Second, we used a finite number of sampling chains with only 100 generations. Future work can explore what would be the effect of changing these hyperparameters. Finally, we only tested participants in the UK. It would be informative to test US participants as well as a wider range of speakers of other languages \citep{BLASI20221153}.

More importantly, it is crucial to acknowledge that employing free elicitation methods could inadvertently generate sentences that reinforce societal biases, including racial and gender stereotypes. However, future research could investigate alternative filtering approaches, possibly involving human moderation, to actively reduce biases within the Sampling with People iterations ~\citep{vanrijn2022bridging}. Nonetheless, our approach can be used with participants in any language, which can help with creating AI systems for low-resource languages~\citep{atari_xue_park_blasi_henrich_2023, rathje_mirea_sucholutsky_marjieh_robertson_vanbavel_2023}. In particular, we are excited about the potential application of our approach to studying cross-cultural differences in tone of voice.  We also believe that our research holds the potential to facilitate nuanced cross-cultural communication and support the development of AI systems that communicate effectively with users from diverse backgrounds. 

\section*{Ethics}
We run our human experiment applying best practices in the responsible and ethical treatment of human subjects. We have deliberately reviewed the ACL Code of Ethics and confirm that our work conforms to all principles addressed in this code. All human experiment participants are paid \$9 per hour and join the experiment after signing an informed consent of an approved protocol. Specifically, participants were recruited online via Prolific and provided consent in accordance with an approved protocol (MaxPlanck Ethics Council \#2021 42).

In our Sampling with People human experiments, to avoid the appearance of profane words or triggering topics, we have added a profanity filter to our experiment such that responses containing profanity or vulgar words cannot be propagated along our sampling chains. Data was collected anonymously (beyond participants' Prolific IDs that were used for compensation). The published data was fully anonymized.

\bibliography{custom}
\clearpage
\newpage
\appendix

\begin{table*}
    \centering
    \begin{tabular}{c||c|c|c}
        Experiment & \# Human Participants & \# Human Responses & \# GPT Responses \\
        \hline
        \hline
        \hline
        SP Sampling & 955 & 9000 & 9000 \\
        Quality-of-fit Rating & 275 & 19470 & 16000 \\
        Similarity Judgment & 71 & 4100 & 4100\\
        Tone Feature Rating & 38 & 800 & 800\\
        \hline
        Overall & 1339 & 33370 & 29900
    \end{tabular}
    \caption{Summary Table. Number of human participants, human judgments, and GPT judgments involved in our paper.}
    \label{tbl:n_participants}
\end{table*}

\begin{figure*}
    \centering
    \begin{subfigure}{\textwidth}
        \centering
        \includegraphics[width=\textwidth]{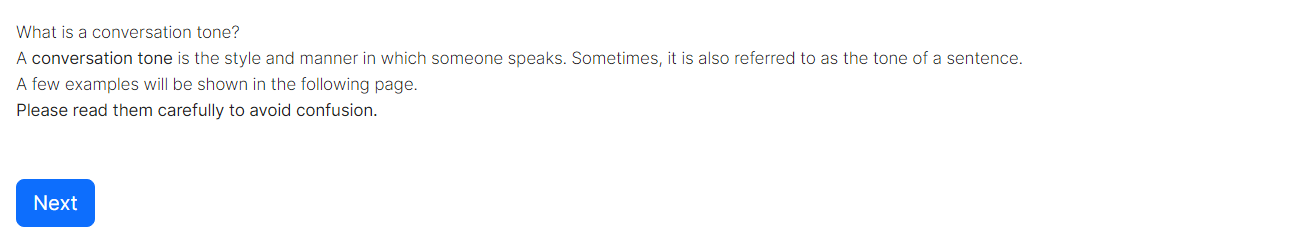}
        \caption{Definition of conversational tones shown to participants.}
        \label{fig:operationalize-a}
    \end{subfigure}
    \begin{subfigure}{\textwidth}
        \includegraphics[width=\textwidth]{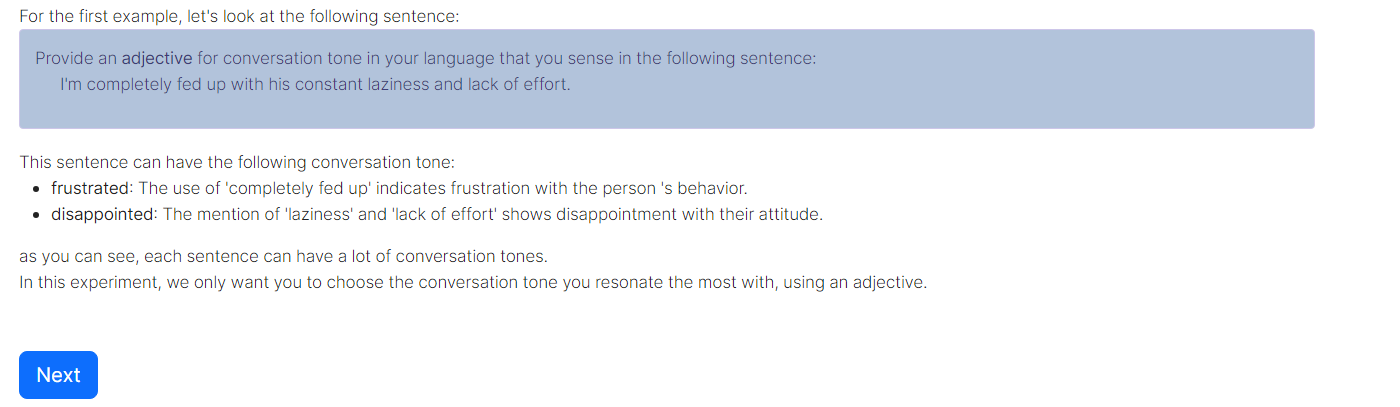}
        \includegraphics[width=\textwidth]{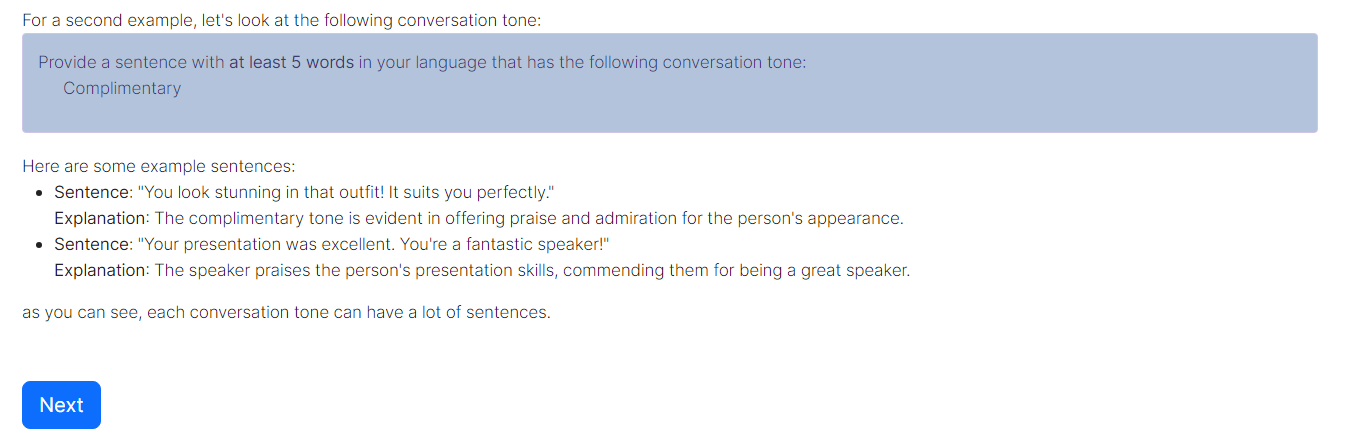}
        \caption{Example scenarios and appropriate example answers.}
        \label{fig:operationalize-b}
    \end{subfigure}
    \begin{subfigure}{\textwidth}
        \includegraphics[width=\textwidth]{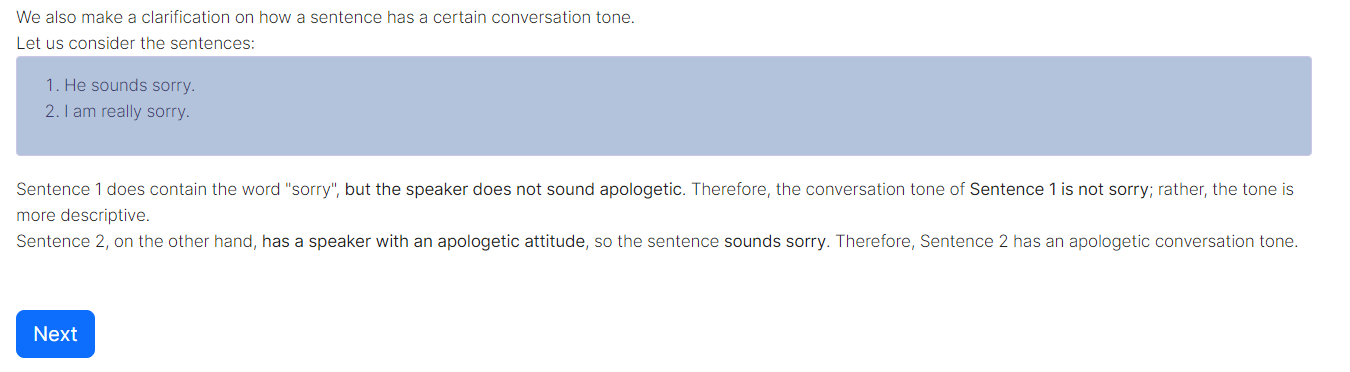}
        \caption{Two examples explaining the concept of conversational tones.}
        \label{fig:operationalize-c}
    \end{subfigure}
    \caption{Instructions shown to participants.}
    \label{fig:operationalize}
\end{figure*}

\begin{figure*}[ht]
    \centering
    \begin{subfigure}{\textwidth}
        \centering
        \begin{small}
            \begin{verbatim}
Create 50 data according to the following format:
{
    ''example_tone'': An adjective that describes a conversational tone,
    ''example_sentence_1'': A sentence with at least 7 words that has the a conversational tone
    of example_tone,
    ''example_sentence_1_explanation'': A 20~30 word explanation on 2~3 conversational tones of
    example sentence 1,
    ''example_sentence_2'': A different sentence with at least 7 words that has the a conversation
    tone of example_tone,
    ''example_sentence_2_explanation'': A 20~30 word explanation on 2~3 conversational tones of
    example sentence 1
}
            \end{verbatim}
        \end{small}
        \caption{ChatGPT prompt for creating seeds of instructional examples on how to create sentences from conversational tones.}
    \end{subfigure}
    
    \vspace{0.3cm}
    \begin{subfigure}{\textwidth}
        \centering
        \begin{small}
            \begin{verbatim}
Create 50 data according to the following format:
{
    ''example_sentence'': A sentence with at least 7 words,
    ''example_tone_1'': An adjective representing a different conversational tone you observe
    from ''example_sentence'',
    ''example_tone_2'': An adjective representing a different conversational tone you observe
    from ''example_sentence'',
    ''example_tone_1_explanation'': Explain how you observed tone 1 from ''example_sentence''
    with 20 to 30 words,
    ''example_tone_2_explanation'': Explain how you observed tone 2 from ''example_sentence''
    with 20 to 30 words
}
            \end{verbatim}
        \end{small}
        \caption{ChatGPT prompt for creating seeds of instructional examples on how to detect conversational tones from sentences.}
    \end{subfigure}
    \caption{ChatGPT prompts for creating seeds that generate the text seen in interface detailed at Figure \ref{fig:operationalize-b}}
    \label{fig:gpt-definition-producer}
\end{figure*}

\begin{figure*}
    \begin{subfigure}{\textwidth}
        \includegraphics[width=\textwidth]{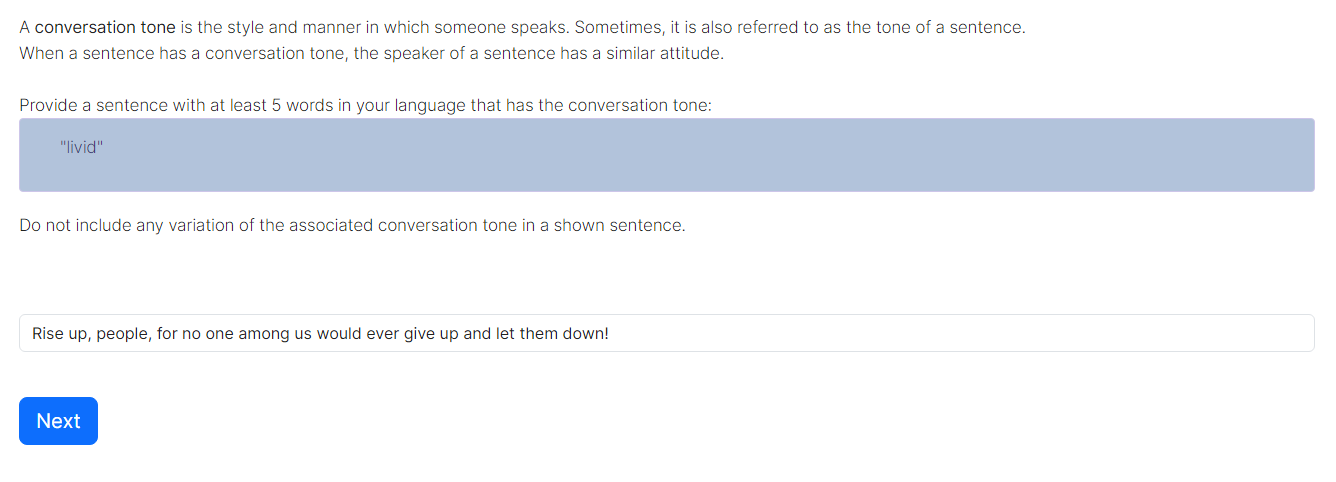}
        \caption{The interface of a S trial.}
        \label{fig:s_trial}
    \end{subfigure}
    \begin{subfigure}{\textwidth}
        \includegraphics[width=\textwidth]{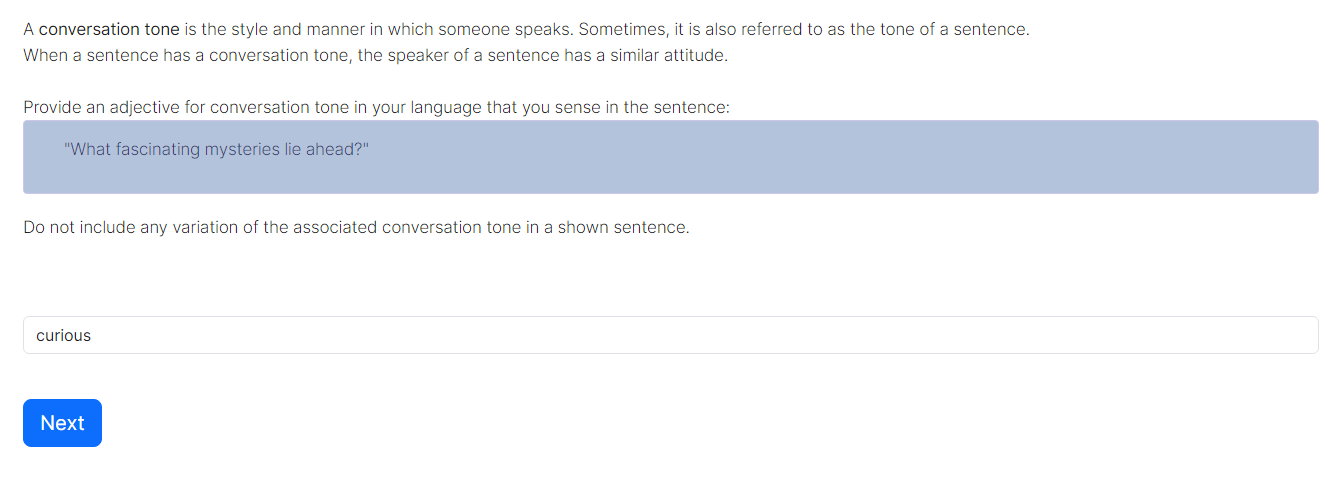}
        \caption{The interface of a T trial.}
        \label{fig:t_trial}
    \end{subfigure}
    \caption{The interfaces of a S trial and a T trial in our Sampling with People human experiment.}
    \label{fig:sp_interface}
\end{figure*}

\section{Appendix: Additional Methods} \label{sec:appendix_methods}
\subsection{Implementation of the Experiments}
\textbf{\textit{Human Experiments.}} Human experiments are implemented using Psynet~\citep{harrison2020gibbs}, a Python package for implementing complex online psychology experiments. PsyNet automates the online hosting of experiments and automatically pays participants through a variety of recruitment platforms, such as Prolific.

\textbf{\textit{GPT Experiments.}} GPT experiments are implemented in Python and GPT responses are all elicited using its chat completion mode. We always used a temperature of 0.8 to elicit responses with higher variance. In all experiments, we used GPT-4 (the June 13th, 2023 release).

\textbf{\textit{Licensed Code}} The implementations of unsupervised alignment methods that we reference for GWOT and latent variable model-based Bilingual Lexicon Induction follow respectively a CC BY-NC 4.0 license and a GPL-3.0 license.

\textbf{\textit{Data Anonymity}} We confirm that all human-originated data provided with our submission are anonymized.

\textbf{\textit{Computational Budget}} We used an AWS server to host our online human experiments. We used an AWS EC2 m5.2xlarge container for 90 hours for completing all necessary experiments once, and 1 AMD Ryzen 7 5800 CPU for 45 hours for all analyses and GPT requests once. No GPUs were used in the progress of this work.

\subsection{Participants and Procedure}
We recruited participants from the crowd-sourcing recruiting service Prolific (\url{https://prolific.com/}).  Participants were compensated approximately 9 GBP per hour for their time.  To reduce cultural differences between English-speaking participants, we target specifically English-speaking participants in one country. All participants satisfied the following three criteria: (1) Were over 18 years of age, (2) Lived in the United Kingdom, (3) Native English speakers. Participants were recruited online via Prolific and provided consent in accordance with an approved protocol (MaxPlanck Ethics Council \#2021 42).

\textbf{\textit{Sampling with People Experiment.}} We recruited 955 human participants for Sampling with People (SP) experiments. These experiments had the following procedure. Each participant completed 10 to 12 trials. For each trial, participants were randomly assigned to either annotate a sentence with a conversational tone (T trials; Figure \ref{fig:schematics-overview}A, \ref{fig:t_trial}) or create a response sentence matching a displayed conversational tone (S trials; Figure \ref{fig:schematics-overview}A, \ref{fig:s_trial}). There were 90 SP sampling chains, each with 100 iterations per chain (50 S and T trials). This results in exactly 4,500 generated sentences and 4,500 tone annotations.

\textbf{\textit{Quality-of-fit Rating Experiment.}} We recruited 275 human participants for the Quality-of-fit Rating experiments. Participants performed 12 rating trials, where they rated each pair of sentence-tones on five Likert scales (1 being the weakest, 5 being the strongest) based on the strength of the conversational tone. For each sentence-tone pair, we collected approximately five ratings. Overall we collected 5 responses for all participants and sentence-tone pairs. 

\textbf{\textit{Similarity Judgement Experiment.}} We recruited 71 human participants for the experiment. Participants performed 50 to 60 trials where they were asked to provide five similarity judgments per pair of (not necessarily distinct) conversational tones, on a scale from 1 to 5, with 1 being the most dissimilar and 5 being the most similar. We then computed the resulting similarity for each conversation-tone pair $(t_1, t_2)$, normalized to the scale of [0, 1].

\textbf{\textit{Tone Feature Rating Experiment.}} We recruited 38 human participants for the experiment.  In this experiment, participants performed 30 to 40 trials where they were asked to rate tone-feature pairs on a 1-to-5 Likert scale. We tested four tone features (positiveness in valence, emotional arousal, informational, relational; see below for definition). The definitions of these features which were provided to participants and GPT agents are listed in Appendix \ref{sec:appendix-methods-feature-def}. The resulting rating for each sentence-tone pair was the average of the obtained five ratings.

\subsection{Explaining the Concept of Conversational Tone}
In all human experiments, participants first receive instructions explaining what a conversational tone is and example sentences clarifying the concepts.

Upon entering the experiments, all participants are presented with this definition of conversational tone. After being presented with an operationalized definition of conversational tone, participants are provided examples that show usages of different conversational tones. Participants will be shown an example regarding detecting a conversational tone from a sentence, and an example of creating a sentence that has a provided conversational tone. These interfaces are exemplified in Figure \ref{fig:operationalize}.

There is a pool of 50 items for each type of example, all created via ChatGPT. To mitigate bias resulting from examples, participants are shown a random item from each pool of examples. The prompt of creating such examples is exhibited as shown in Figure \ref{fig:gpt-definition-producer}.

\subsubsection{Consideration for human instructions and GPT prompts}
In pilot experiments, several participants create sentences of a provided conversational tone $T$ in the form of ``<Subject> felt <T>''. However, such a sentence may not necessarily convey conversational tone $T$. For example, for the conversational tone ``apologetic'', the sentence ``He felt apologetic'' communicates that a ``He'' is apologetic, but not that the speaker is apologetic, which contradicts the definition of conversational tone as ``the speaker's attitude in a conversation''. So, as an effort to prevent participants from providing such responses, an instruction as listed below is presented as in Figure \ref{fig:operationalize-c}.

\begin{figure*}
    \centering
    \includegraphics[width=\textwidth]{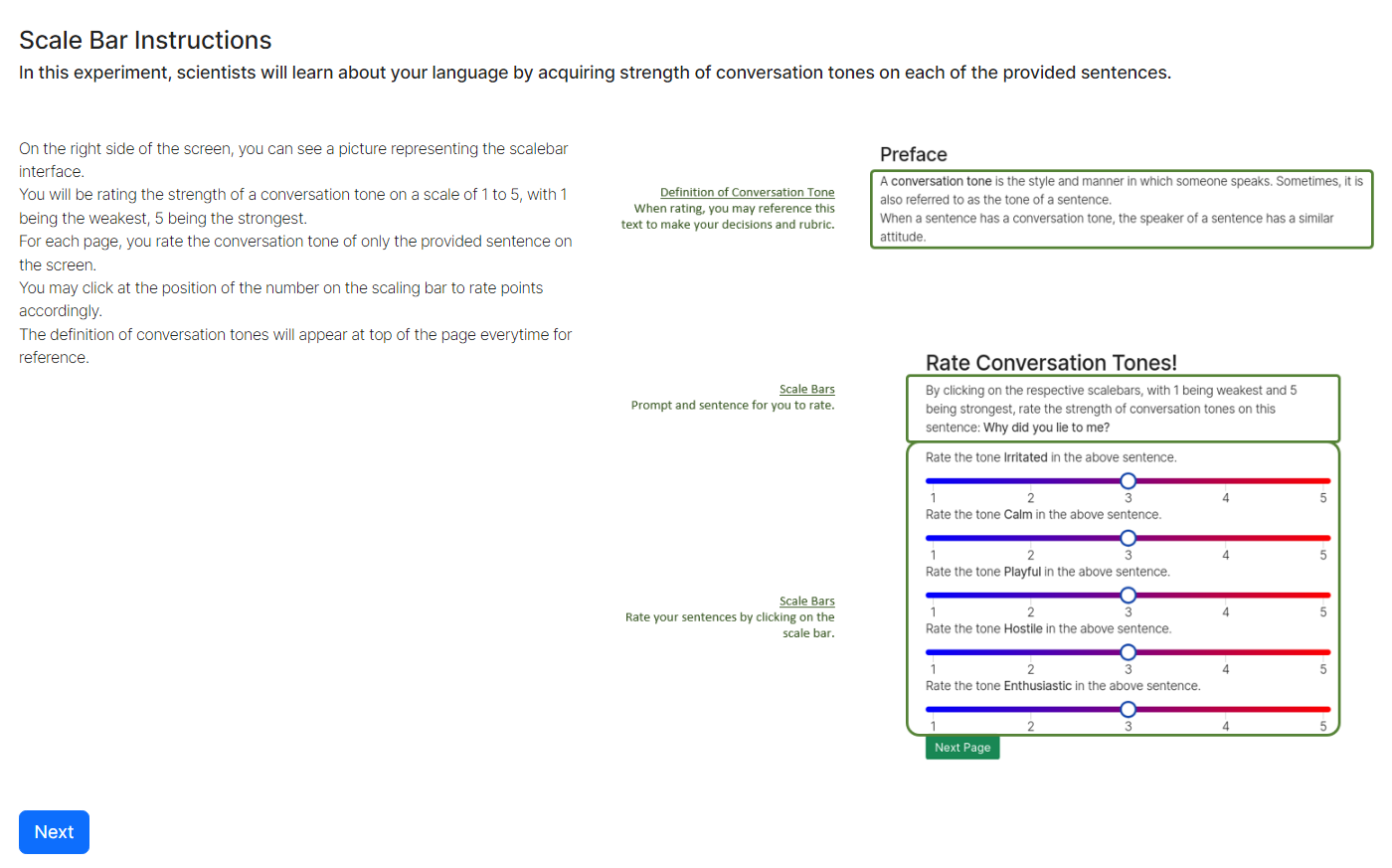}
    \caption{This page of the experiment details the instructions for using our designed scale interface for rating.}
    \label{fig:define-scaling}
\end{figure*}

\subsubsection{Slider Instruction}
In some experiments, participants will need to rate some properties of conversational tones on a Likert scale of 1 to 5. For those experiments, after explaining the concept of conversational tones, they are presented with the following introduction to learn how to use the slider used for rating as depicted in Figure \ref{fig:define-scaling}.

\subsection{Experiment 1: SP Sampling} \label{sec:appendix-methods-sp}
\subsubsection{Human Experiment}
This section describes the implementational details of SP Sampling (see Section \ref{sec:methods}) human and GPT experiments.

The participants first go through the general instructions (see Figure \ref{fig:operationalize}) and do two practice trials to familiarize themselves with the tasks they will be performing: (1) detecting a conversational tone from a sentence and (2) creating a sentence that conveys some provided conversational tone.

In the main experiment, each participant does ten trials. The human interface is shown in Figure \ref{fig:sp_interface}

\begin{table*}
    \centering
    \begin{tabular}{c||c|c}
        Response Type & Validation Criterion & Implementation \\
        \hline
        Sentence & Must have more than 5 words & \texttt{RegEx} \\
        Sentence & Must be a grammatically correct sentence & \texttt{GingerIt} \\
        Tone & The response can only contain alphabets and hyphens & \texttt{RegEx} \\
        Tone & The response must be an adjective & \texttt{PyDictionary} \\
        Tone & The response must be correctly spelled & \texttt{PyDictionary} \\
        Both & Cannot contain any stemmed variation of prompt & \texttt{nltk} \\
        Both & Cannot contain profanity & \texttt{profanity-check}
    \end{tabular}
    \caption{Summary of Sampling with People response filters}
    \label{tbl:sp-filters}
\end{table*}

To avoid low-quality sentences, we automatically check the submitted sentence for the following criteria: (1) The sentence has to have more than five words; (2) The sentence is grammatically correct (checked using \texttt{gingerit} \footnote{\texttt{gingerit}: \url{https://github.com/Azd325/gingerit/blob/main/gingerit/gingerit.py}}) (3) The sentence does not contain any stemmed variation of some word that is already in the provided conversational tone (for example ``politely'' for ``polite'') using \texttt{nltk}\footnote{https://www.nltk.org/}; (4) Does not contain offensive or vulgar words checked with the Python package \texttt{profanity-check} \footnote{\texttt{profanity-check}: \url{https://pypi.org/project/profanity-check/}}.

The conversational tones were verified in a similar fashion: (1) The tone response did not contain any stemmed variation of some word that is already in the provided conversational tone or sentence, (2) The response must be an adjective using \texttt{PyDictionary}\footnote{https://pypi.org/project/PyDictionary/}, and (3) It must not contain profanity. See Table \ref{tbl:sp-filters} for a summary.

In the human experiment, we elicited 90 sampling chains, each with 100 iterations. Participants cannot revisit the same chain.

\subsubsection{GPT-4 Experiments}
The GPT prompts (see Figure \ref{fig:gpt_prompt_sre_sampling})  were nearly identical to the human experiments. In the GPT instance, we similarly elicited 90 sampling chains, each with 100 iterations.

\subsection{Experiment 2: SP Quality-of-fit Rating}
\begin{figure*}
    \centering
    \begin{subfigure}{\textwidth}
        \centering
        \begin{small}
            \begin{verbatim}
A conversational tone is the style and manner in which someone speaks.
Provide an adjective for conversational tone in English that you sense in the following sentence:
<sentence of previous response>. Respond using only an adjective.
            \end{verbatim}
        \end{small}
        \caption{GPT Prompt for sampling a conversational tone given a sentence.}
    \end{subfigure}
    
    \vspace{0.3cm}
    \begin{subfigure}{\textwidth}
        \centering
        \begin{small}
            \begin{verbatim}
A conversational tone is is the style and manner in which someone speaks.
Provide one sentence with at least five words in English that has the conversational tone:
<conversational tone of previous response>.
            \end{verbatim}
        \end{small}
        \caption{GPT Prompt for sampling a sentence given a conversational tone.}
    \end{subfigure}
    \caption{Collection of GPT Prompts for GPT participant in sampling and rating aspects of SP paradigm.}
    \label{fig:gpt_prompt_sre_sampling}
\end{figure*}

\begin{figure*}
    \centering
    \includegraphics[width=\textwidth]{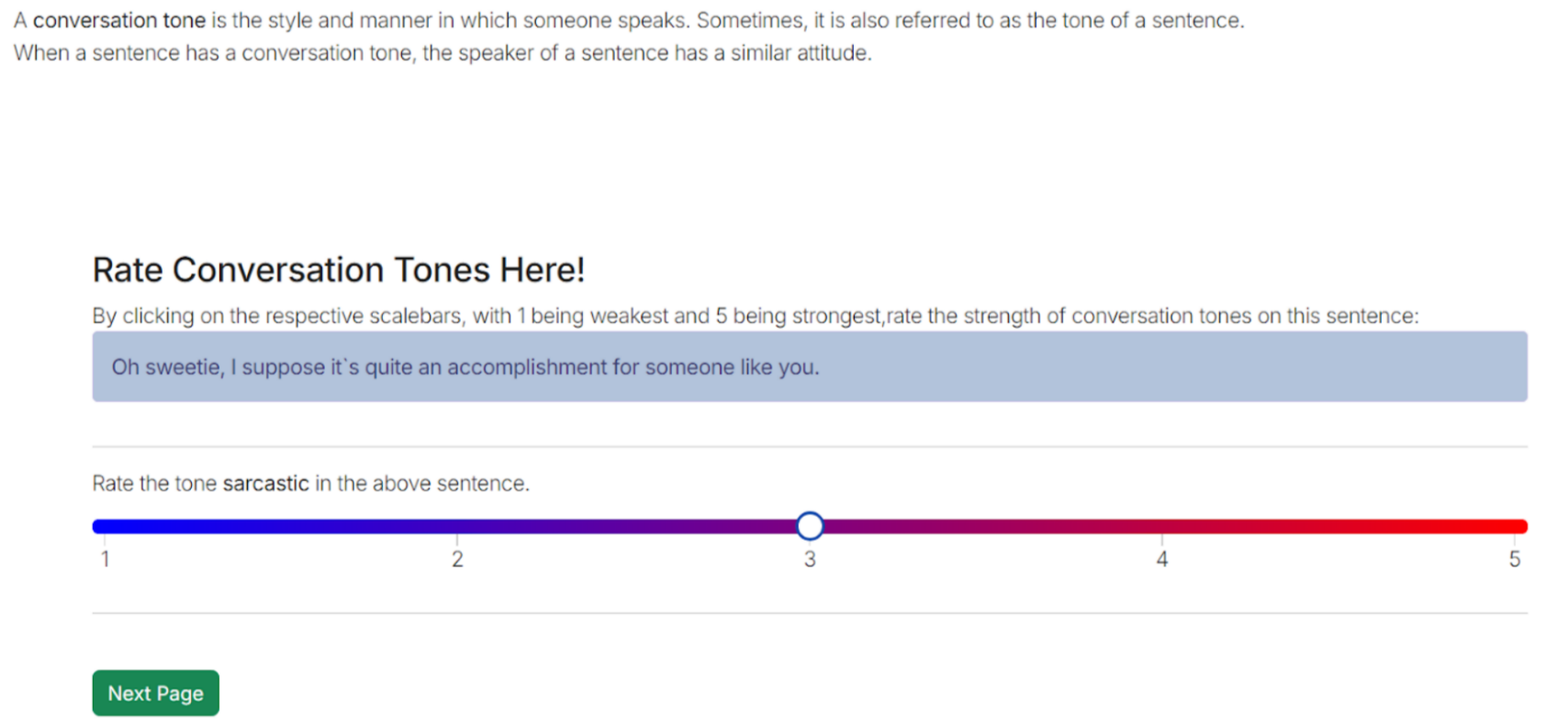}
    \caption{Example interface for quality-of-fit rating human experiment interface.}
    \label{fig:dr_interface}
\end{figure*}

\begin{figure*}
    \centering
    \begin{subfigure}{\textwidth}
        \centering
        \begin{small}
            \begin{verbatim}
A conversational tone is the style and manner in which someone speaks.
On a scale of 1 to 5, with 5 being strongest, how strong is the provided conversational tone in
the following English sentence?
Tone: <conversational tone to rate>
Sentence: <sentence to rate>
Respond with only a number.
            \end{verbatim}
        \end{small}
        \caption{GPT Prompt for rating the strength of a conversational tone on a sentence.}
    \end{subfigure}
    \caption{GPT Prompts used for quality-of-fit rating experiment.}
    \label{fig:gpt_prompt_sre_rating}
\end{figure*}

\subsubsection{Human Experiment}
After reading the general instruction (see Figure \ref{fig:operationalize}), participants first do two practice trials to familiarize themselves with the task they will be performing: rating the strength of a conversational tone when given a sentence. Then, participants proceed to the main experiment. Participants can only rate a conversational tone-sentence once, and each tone-sentence pair is rated by approximately 5 distinct participants. The strength of conversational tones in a sentence is rated on a Likert scale from 1 to 5, with 1 being the weakest and 5 being the strongest. The final rating of such a tone-sentence pair would be the average of all 5 ratings. An example of the rating interface is provided in Figure \ref{fig:dr_interface}.

\subsubsection{GPT-4 Experiments.}
GPT receives the prompt for quality-of-fit rating as outlined in Figure \ref{fig:gpt_prompt_sre_rating}.

\subsection{Experiment 3: conversational tone Similarity Judgment}
\begin{figure*}
    \centering
    \begin{subfigure}{\textwidth}
        \centering
        \includegraphics[width=\textwidth]{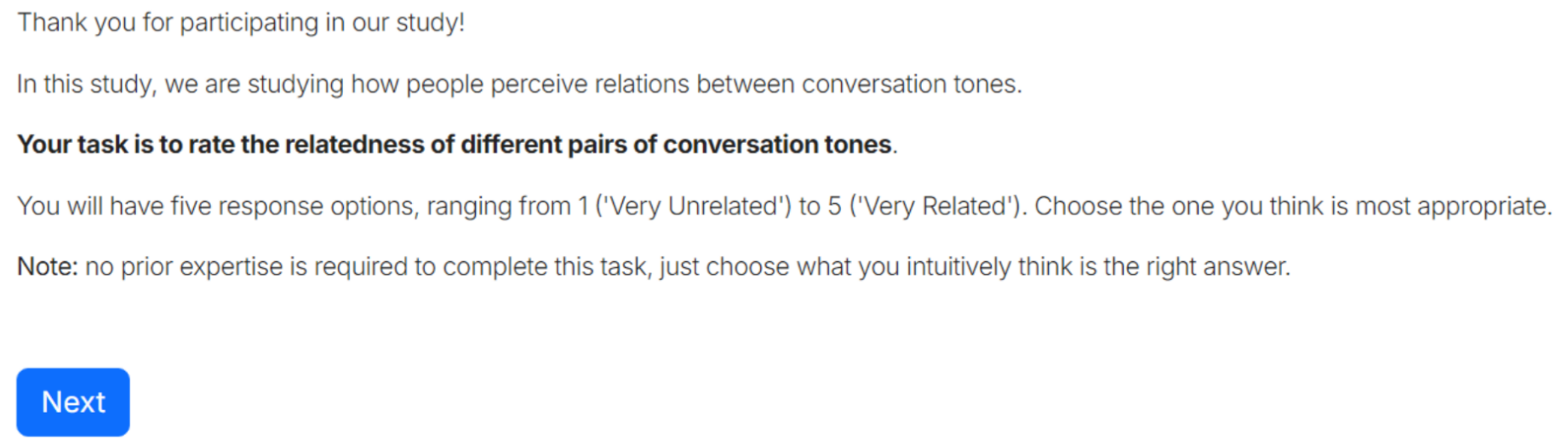}
        \caption{Interface that introduces human participants to similarity judgment experiment.}
        \label{fig:sjt_intro}
    \end{subfigure}
    \begin{subfigure}{\textwidth}
        \centering
        \includegraphics[width=\textwidth]{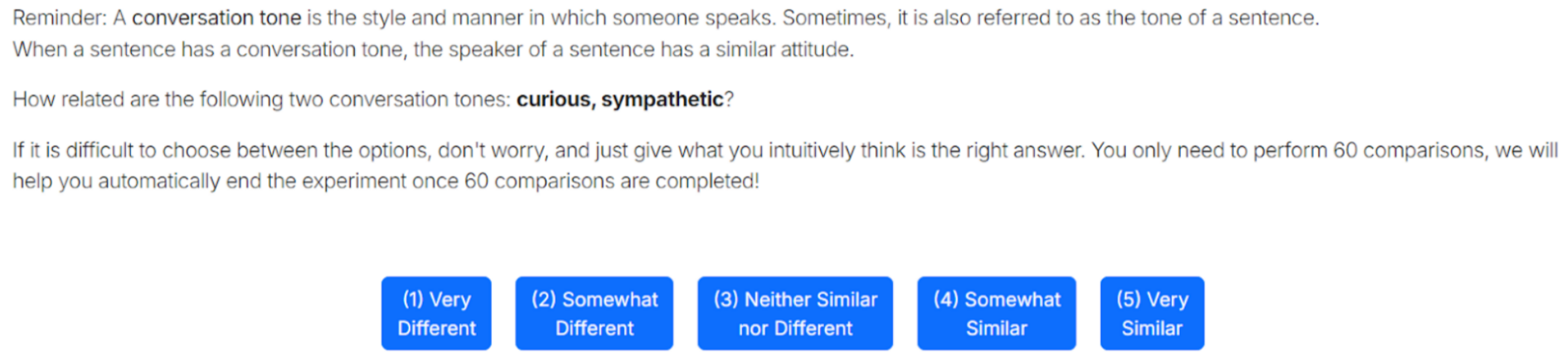}
        \caption{Interface that represents the main body of similarity judgment human experiment.}
        \label{fig:sjt_body}
    \end{subfigure}
\end{figure*}

\begin{figure*}
    \centering
    \begin{subfigure}{\textwidth}
        \centering
        \begin{small}
            \begin{verbatim}
People described conversational tones using words.
A conversational tone is the style and manner in which someone speaks,
and sometimes, it is also referred to as the tone of a sentence.
How similar are the conversational tones in each pair on a scale of 0-1
where 0 is completely dissimilar and 1 is completely similar?
        conversational tone 1: {tone_a}
        conversational tone 2: {tone_b}
Respond only with the numerical similarity rating.
            \end{verbatim}
        \end{small}
    \end{subfigure}
    \caption{GPT Prompt for rating the similarity of conversational tones.}
    \label{fig:gpt_prompt_sjt}
\end{figure*}
\subsubsection{Human Experiment}
After reading the general (see Figure \ref{fig:operationalize}) and experiment-specific instructions (see Figure \ref{fig:sjt_intro}), participants proceed to the main experiment in which they judge the similarity of a pair of two conversational tones. 

Each participant would rate the similarity of several distinct pairs of conversational tones on a Likert scale of 1 to 5, where 1 represents ``Semantically dissimilar conversational tones'' and 5 represents ``Semantically similar conversational tones''. Each pair of conversational tones would be rated five times, with the average score of those 5 ratings as the final similarity score for such pair. Only distinct pairs are rated. That means the similarity judgment of 40 conversational tones concerns only the 820 distinct pairs among all possible tuples of tones. And, similar to previous experiments, participants cannot rate any conversational tone pair more than once. An example of the rating interface follows in Figure \ref{fig:sjt_body}.

\subsubsection{GPT Experiment}
The prompt for similarity judgment that GPT receives is as outlined in \ref{fig:gpt_prompt_sjt}.

\subsection{Experiment 4: conversational tone Feature Rating} \label{sec:appendix-methods-feature-def}
\begin{figure*}
    \centering
    \begin{subfigure}{\textwidth}
        \centering
        \begin{small}
            \begin{verbatim}
A conversational tone is the style and manner in which someone speaks.
{explanation of rated tone feature using its definition in Table}
On a scale of 1 to 5, where 5 means strongest and 1 means weakest, how {feature} is the conversational tone
''{tone}''?
Respond with only a number.
            \end{verbatim}
        \end{small}
    \end{subfigure}
    \caption{GPT Prompt for rating the strength of features in conversational tones.}
    \label{fig:gpt_prompt_tfr}
\end{figure*}

\begin{figure*}
    \centering
    \includegraphics[width=\linewidth]{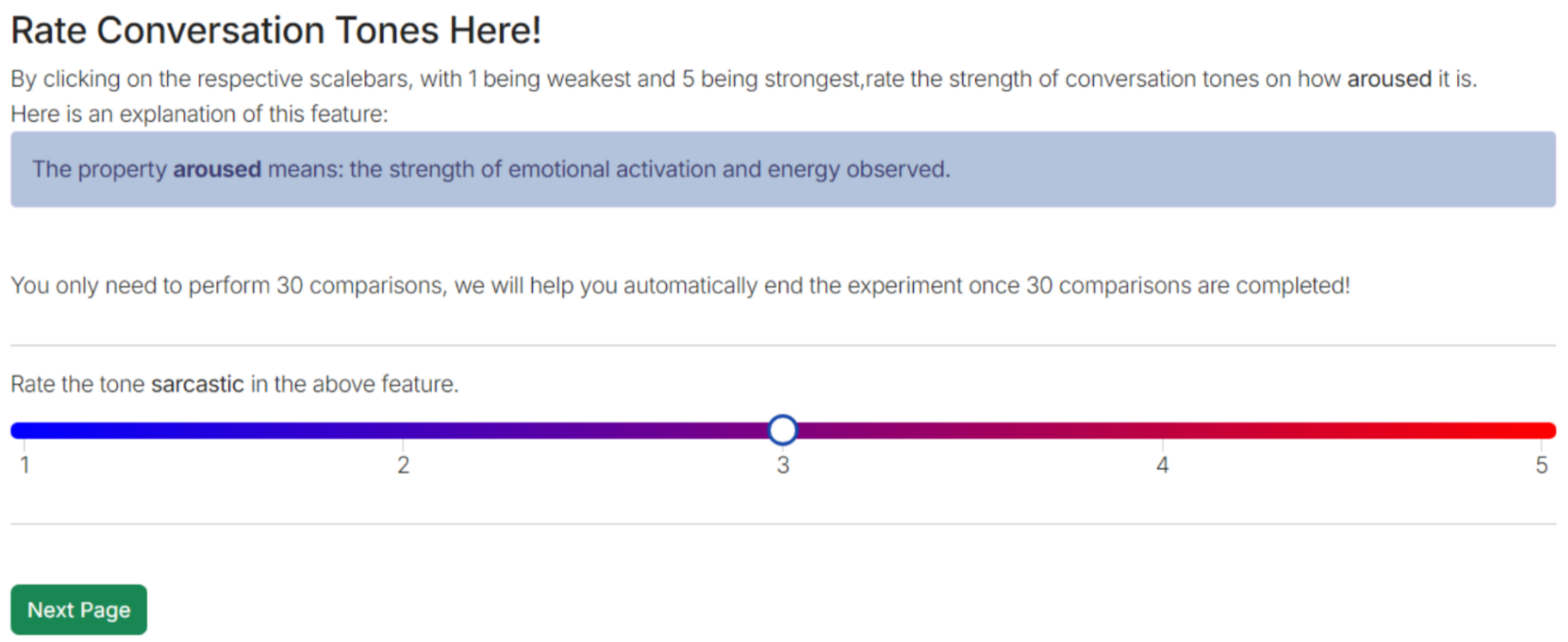}
    \caption{Interface for the main body of tone feature rating human experiment.}
    \label{fig:tfr_interface}
\end{figure*}
\subsubsection{Human Experiment}
After reading the general instructions (see Figure \ref{fig:operationalize}), human participants proceed to the main experiment in which they rate a feature of conversational tones. Based on psycholinguistic literature~\citep{YEOMANS2022293, VonFintel2006-VONMAL, 10.1162/coli_a_00426, facework, portner2009modality} we selected the following features along with their definitions:
\begin{itemize}
    \item \textbf{positive in valence}: Positiveness in valence means the positiveness of emotional valence.
    \item \textbf{aroused}: Aroused means the amount of emotional arousal observed.
    \item \textbf{Informational}: Informational means the extent to which a speaker's motive focuses on giving and/or receiving accurate information.
    \item \textbf{Relational}: Relational means the extent to which a speaker's motive focuses on building the relationship.
\end{itemize}
The strength of features in a conversational tone is rated on a Likert scale from 1 to 5, with 1 being the weakest and 5 being the strongest. Participants are provided an interface for rating the features in conversational tones. See Figure \ref{fig:tfr_interface} for a screenshot of the task.

\begin{figure*}[!ht]
    \centering
    \includegraphics[trim={0 6cm 0 0},clip,width=\textwidth]{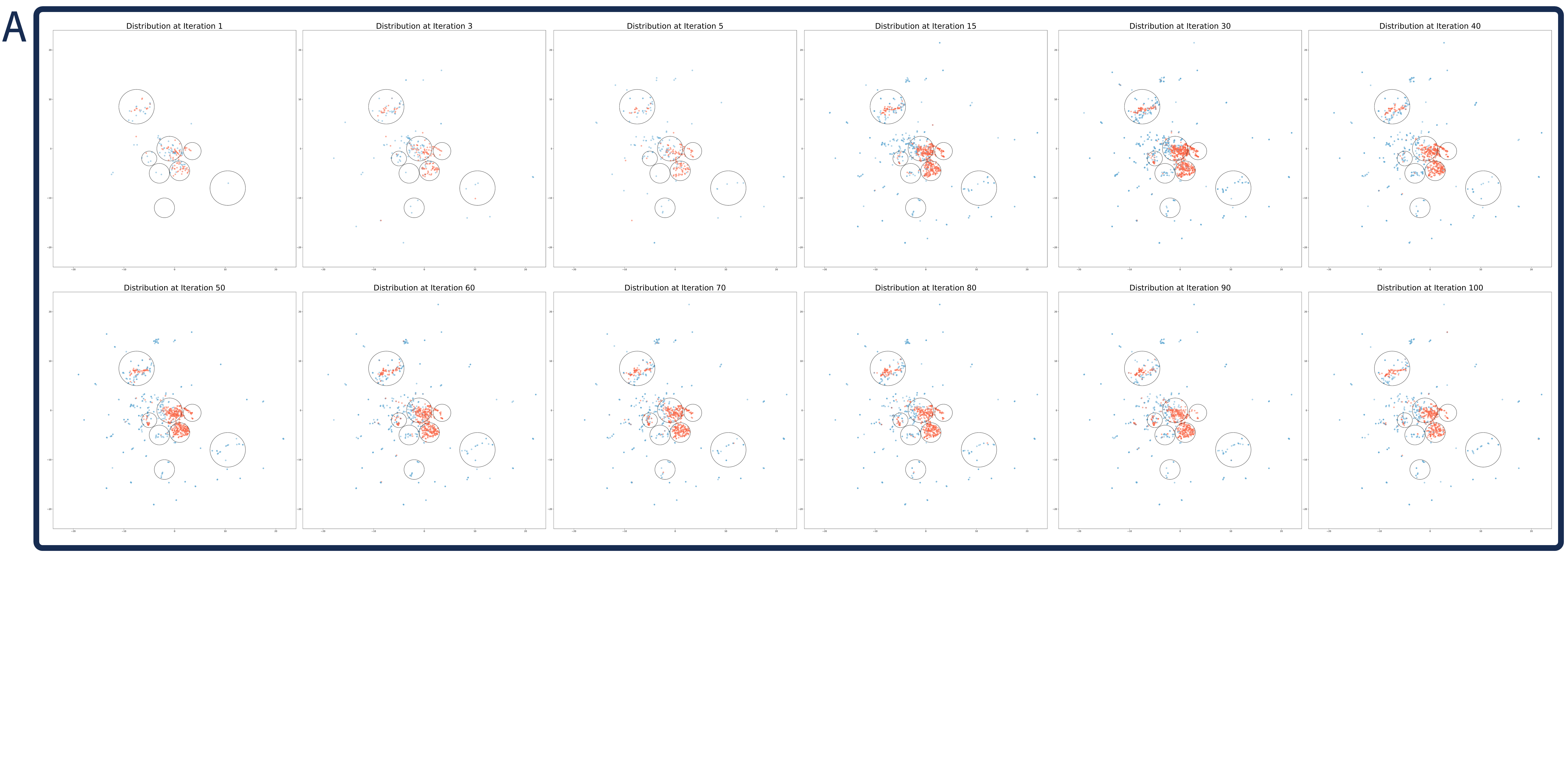}
    \includegraphics[trim={0 6cm 0 0},clip,width=\textwidth]{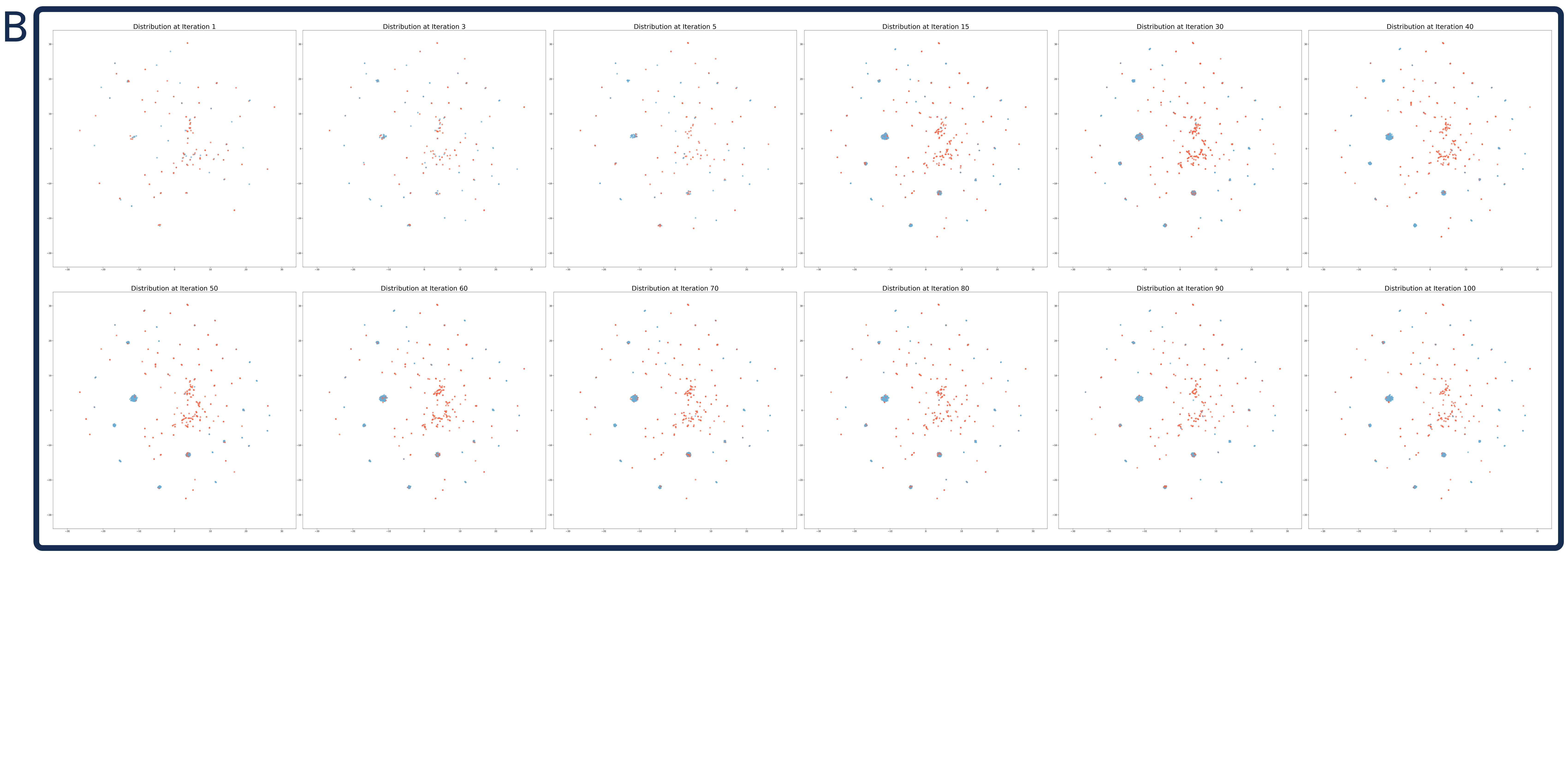}
    \caption{The dynamics of joint-embedding space for sentences and conversational tones throughout a selection of iterations. Embeddings are produced via first obtaining sentence-embeddings or tone word-embeddings using \texttt{distilbert-base-uncased}~\citep{Sanh2019DistilBERTAD} pretrained weights, and then projected onto a 2-dimensional space using \texttt{UMAP} manifold embedding~\citep{mcinnes2020umap}. \textbf{A}: The time evolution of joint-embedding sentence space. \textbf{B}: The time evolution of joint-embedding tone space.
}
    \label{fig:supp-joint-space}
\end{figure*}

\subsubsection{GPT Experiment}
The prompt for tone feature rating that GPT receives is as outlined in \ref{fig:gpt_prompt_tfr}.

\section{Supplementary Statistical Analyses}
\subsection{SP Sampling}

\begin{figure*}
    \centering
    \includegraphics[scale=0.41]{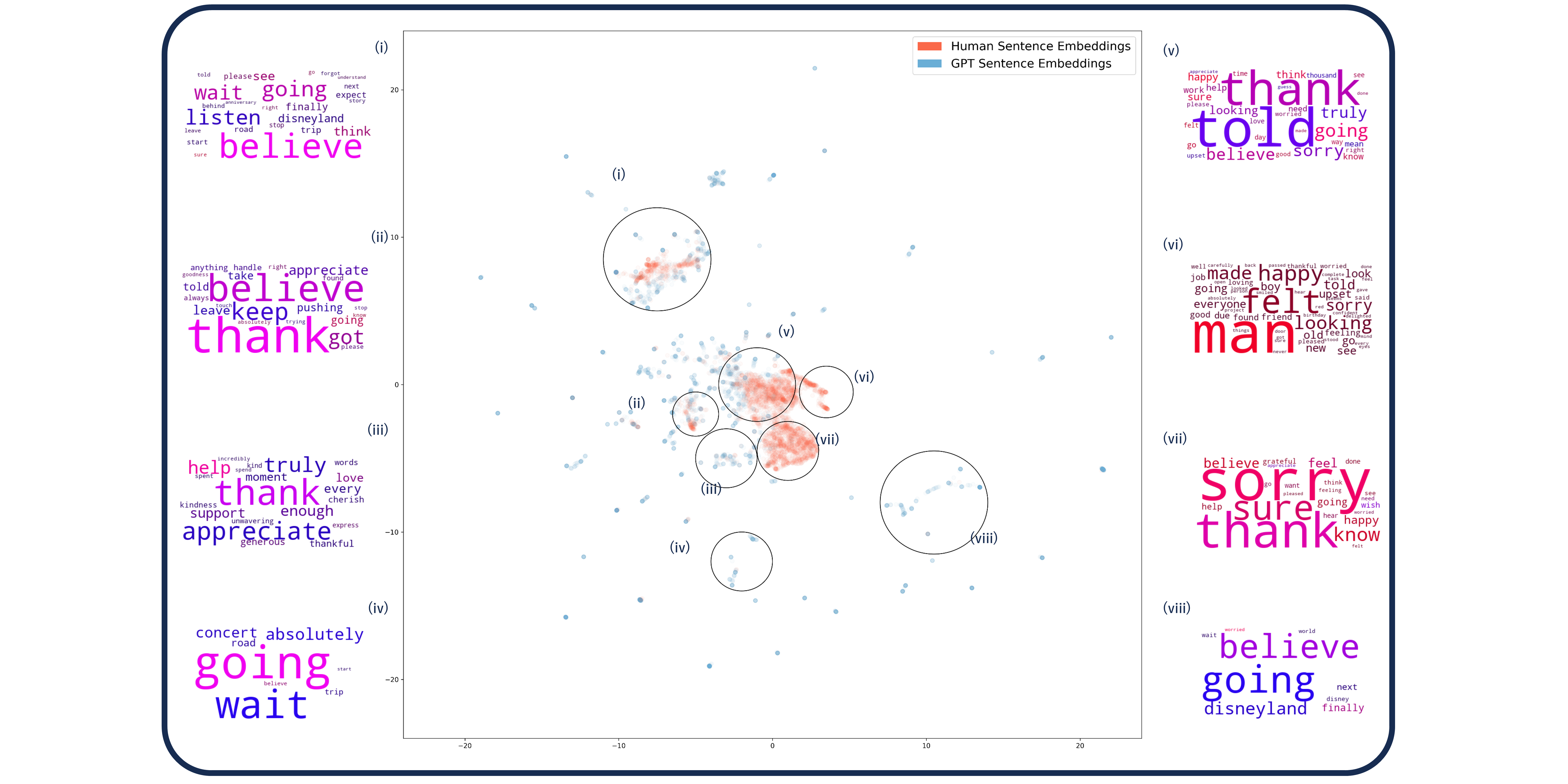}
    \caption{Sentence embedding. Word clouds show the frequency of words (right and left insets) in corresponding circles on the sentence UMAP embedding space (center). Red points resemble each sentence sampled from human instances, and blue points resemble GPT instance sentences. Brighter red and blue hues indicate respectively high TF-IDF~\citep{tfidf} scores in human, GPT sentences in each word cloud (i.e., bright purple words are highly frequent across humans and GPT).}
    \label{fig:SP_wordcloud}
\end{figure*}

\begin{table*}[t]
       \centering
       \begin{tabular}{c|c||c|c|c}
       Performance Category & k & Procrustes & GWOT & BLI\\
       \hline
       \hline
       \hline Domain Similarity & N/A &
       0.625 [0.607, 0.642] &
       0.625 [0.607, 0.642] &
       \textbf{0.8} [0.8, 0.8] \\
       Preservation (Human) & & & & \\
       \hline Domain Similarity & N/A &
       0.499 [0.473, 0.513] &
       0.454 [0.405, 0.519] &
       \textbf{0.82} [0.82, 0.82] \\
       Preservation (GPT) & & & & \\
       \hline
       kNN Matching Rate & $1$ &
       0.392 [0.325, 0.463] &
       0.339 [0.293, 0.388] &
       \textbf{0.657} [0.638, 0.675]  \\
       w.r.t. CC Alignment & $2$ &
       0.41 [0.363, 0.460] &
       0.354 [0.288, 0.397] &
       \textbf{0.635} [0.606, 0.65] \\
       & $3$ &
       0.456 [0.408, 0.496] &
       0.403 [0.350, 0.461] &
       \textbf{0.653} [0.635, 0.667]  \\
       & $4$ &
       0.499 [0.451, 0.541] &
       0.461 [0.397, 0.520] &
       \textbf{0.705} [0.686, 0.719] \\
       & $5$ &
       0.531 [0.476, 0.568] &
       0.503 [0.422, 0.563] &
       \textbf{0.730} [0.709, 0.745] \\ 
       \end{tabular}
   \caption{Table of benchmarking results on proposed metrics for unsupervised cross-domain alignment methods. Procrustes: Orthogonal Procrustes Transformation. GWOT: Gromov-Wasserstein Optimal Transport~\citep{grave2018unsupervised}. BLI: Bilingual Lexicon Induction via Latent Variable Model~\citep{ruder2018discriminative}. Results are aggregated across $100$ seeds for stochastic methods.}
    \label{tbl:benchmark}
\end{table*}

\textbf{\textit{Sample Reliability.}} For testing the reliability of our elicited conversational tone distribution, for both human and GPT instances, we measure the split-half correlation of their conversational tone distributions from the acquired dataset of their SP instances. This split-half correlation is computed along the following procedure. First, we randomly partition a set of SP-gathered data into two halves. Then, we find the frequency of each conversational tone within the dataset's halves. At last, we compute the correlation between the frequency of conversational tones to be the split-half correlation of conversational tone distribution within an SP instance.

Over \textit{N} = 5000 bootstrap processes for both the human and GPT instances, we measure the human conversational tone distribution split-half correlations to be \textit{r} = 0.91 [0.87, 0.93], and the GPT conversational tone distribution split-half correlations to be \textit{r} = 0.87 [0.73, 0.94].

\textbf{\textit{Semantic Interpretation of Sentence Space}}
Figure \ref{fig:SP_wordcloud} shows the joint-embedding space via UMAP~\citep{mcinnes2020umap} for sentences encoded using \texttt{
distilbert-base-uncased} embeddings~\citep{Sanh2019DistilBERTAD} from both humans and GPT. Consistent with our findings regarding tones in the Result section, the distribution was much more concentrated (entropy of 5.05 bits [5.03, 5.07] via bootstrapping) compared with humans (entropy of 4.12 bits [4.10,4.15]). Figure \ref{fig:SP_wordcloud} also shows different topics in different parts of the space, which also shows differences in the produced sentences for humans and GPT. From this figure, we observe high repetition of sentence literal content across many GPT-occupied locations of the shared sentence embedding space (e.g., excited to go to Disneyland), while in regions dominated by humans, we usually observe a higher variance of words used. The highlighted regions in Figure \ref{fig:SP_wordcloud} show the sentence space shows dense semantic topics, (e.g., ``gratefulness'' in circle (ii)).

\subsection{Quality-of-fit Rating Experiment}
\textbf{\textit{Sample Reliability of Correlation Matrices.}} We measure the sample reliability of human perception's and GPT perception's correlation matrix using the following procedure. First, for a set of gathered quality-of-fit ratings, we randomly partition such dataset by sentences. Then, within each partition, we produce a correlation matrix. Finally, we compute the correlation of these matrices (treated as vectors).  We used 5000 bootstrapped dataset. For humans ratings, we find this correlation to be \textit{r} = 0.95 [0.92, 0.96]; for GPT ratings, we find this correlation to be \textit{r} = 0.90 [0.84, 0.93]. The cross-domain matrix itself has a halfsplit correlation of \textit{r} = 0.95 [0.92, 0.96].

\textbf{\textit{Sample Reliability of Similarity Judgment Experiment.}} First, for a set of gathered quality-of-fit ratings, we partitioned the ratings of each conversational tone into two halves and computed the quality-of-fit rating correlation matrix from each half of the quality-of-fit rating data. We then compute the correlation between these similarity matrices. We used 5000 bootstraps samples and found the split-half correlation of human's similarity matrix to be \textit{r} = 0.72 CI = [0.81, 0.84], and that for GPT's similarity matrix to be \textit{r} = 0.987 CI = [0.978, 0.983].

\subsection{Cross-Domain (Cross-Correlation) Analysis} \label{sec:appendix-cca}
\textbf{\textit{Computation of Feature Arrowmarks.}}
We used linear regression to regress the tone rating for each of the four theoretic dimensions using a projection technique and the responses of this experiment (MDS biplot~\citep{greenacre2010biplots} treating the tone feature ratings as biplot arrows as shown in Figure \ref{fig:mds}A. The concrete computation process is as follows. First, we construct a feature rating vector $\vec{f}_i$ for each vector. Then, we fit these features and the MDS embedding $x$ coordinates ($\vec{x}$) using linear regression, arriving at some regression line:
$\hat{\vec{x}} = \sum_i \alpha_i \vec{f}_i$
The coefficient $\alpha_i$ is then taken to be the $x$-axis direction of feature $i$'s arrowmark. The same procedure was performed to compute the arrowmarks' $y$-axis direction. Note that when computing the arrowmark for humans' feature rating, we only fit the feature ratings to the humans' conversational tone MDS solution. GPT's arrowmark dimensions was only fitted to GPT's feature ratings too.

\textbf{\textit{Cosine Similarity of Features.}} As performed in Section \ref{sec:result-cca}, we bootstrap over the cosine similarity of these feature vectors over different MDS solutions, and find that while the feature ``informational'' is consistently aligned with high cosine similarity in arrowmark direction across both groups (mean 0.98 CI = [0.97, 0.99]), the feature ``relational'' is not so strongly aligned (mean 0.6 CI = [0.57, 0.87]). Furthermore, the features ``positive in valence'' and ``aroused'' both observe negative cosine similarity in directions (respectively, mean -0.69 CI = [-0.71, -0.64]; mean -0.65 CI = [-0.69, -0.4]). This suggests a deviation between the human and GPT understanding of these features.

\textbf{\textit{Explained Variance of Features.}} Additionally, we investigate the significance of each feature vector by computing its ``explained variance'' within the shared embedding space. We compute the explained variance of a feature vector is computed as the variance of scalar projections of all MDS tone embeddings onto that feature vector. For GPT tone embeddings, the order of conversational tone features from highest to lowest explained variance is ``positive in valence'' (mean 0.873 CI = [0.872, 0.878]), ``relational'' (mean 0.45 CI = [0.42, 0.76]), ``aroused'' (mean 0.17 CI = [0.16, 0.24]), then ``informational'' (mean 0.126 CI = [0.12, 0.127]). For human tone embeddings, the order of features from highest to lowest explained variance is instead ``positive in valence'' (mean 0.71 CI = [0.69, 0.87]), ``aroused'' (mean 0.54 CI = [0.5, 0.79]), ``relational'' (mean 0.37 CI = [0.34, 0.76]), followed by ``informational'' (mean 0.3 CI = [0.27, 0.79]). In both spaces, we find ``positive in valence'' to be a dominant dimension of conversational tone embeddings, while humans and GPT do not fully agree upon the dominance of other directions.

\section{Hyperparameters in Alignment Paradigms}

\textbf{\textit{Gromov-Wasserstein Optimal Transport (GWOT).}}
For GWOT, we adopted Grave et al.'s implementation~\citep{grave2018unsupervised, conneau2017word, lample2017unsupervised} using 500 iterations for its convex initiation, and a learning rate of 10, batch size of 10, regularization coefficient of 0.5, with 15 epochs for its stochastic iteration in GWOT procedure.

\textbf{\textit{Bilingual Lexicon Induction via Latent Variable Model.}}
For this method, we adopted Ruder et al.'s implementation~\citep{ruder2018discriminative, artetxe2016emnlp, artetxe2017acl, artetxe2018aaai, artetxe2018acl}. During lexicon induction, we used a backward direction, considering 5 nearest neighbors in translation retrieval. We did not use a seed dictionary. We also made small modifications to Ruder et al.'s implementation (model training batch size from 1000 to 5) to adapt the paradigm towards our smaller set of embeddings.

\section{Declaration of Generative AI and AI-Assisted Technologies in the Writing Process}
During the preparation of this work, we sometimes used GPT for edits. After using this tool, the authors reviewed and significantly edited the content as needed and took full responsibility for the content of the publication. Additionally, we used \textit{word-tune} \url{https://www.wordtune.com/} and \textit{Grammarly} (\url{https://www.grammarly.com/}) to check syntax and proofread the document. Writing the experimental code we used code-suggestions by \textit{Microsoft Copilot} (\url{(https://copilot.microsoft.com/}). We reviewed all suggestions to make sure they reflected our intentions.

\section{Enlarged Figures from Main Paper}
\begin{figure*}[t]
    \centering
    \includegraphics[width=1.2\textwidth,angle=90]{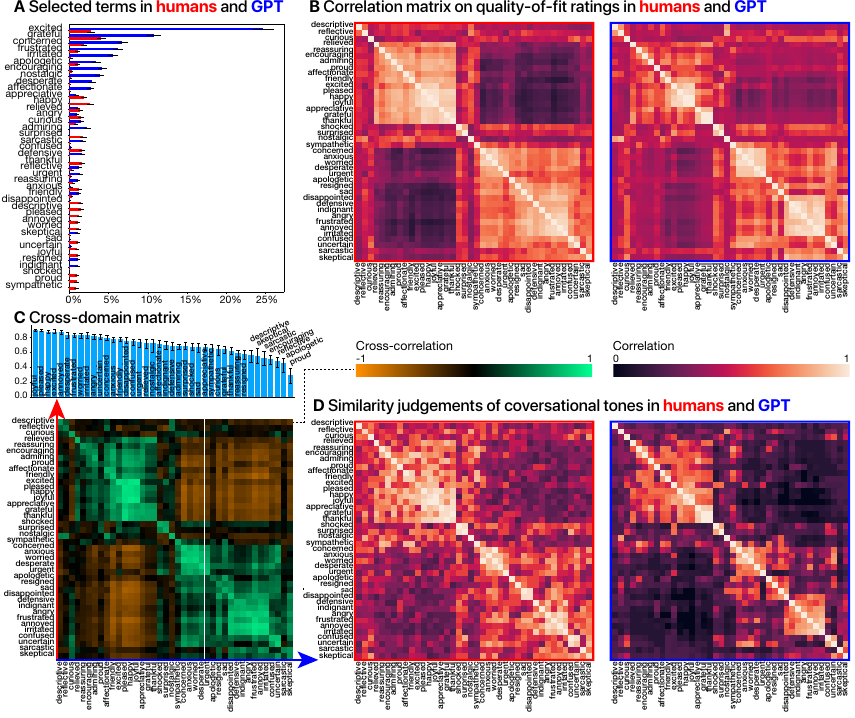}
    \caption{Enlarged version of Figure \ref{fig:dense}}
    \label{fig:dense-large}
\end{figure*}

\begin{figure*}[t]
    \centering
    \includegraphics[width=1.5\textwidth,angle=90]{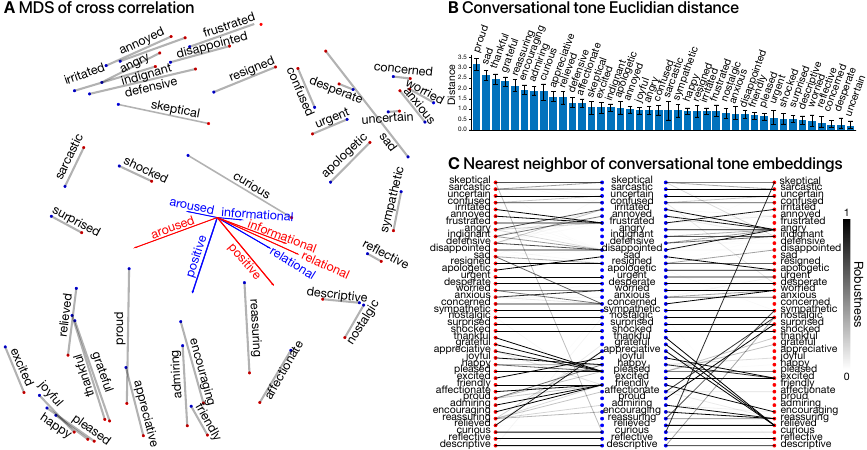}
    \caption{Enlarged version of Figure \ref{fig:mds}}
    \label{fig:mds-large}
\end{figure*}

In this section, we attach enlarged Figures \ref{fig:dense}, \ref{fig:mds} as Figure \ref{fig:dense-large}, \ref{fig:mds-large} from the main paper for readability.

\end{document}